\documentclass{article}

\usepackage{iclr2027_conference,times}

\usepackage{microtype}
\usepackage{graphicx}
\usepackage{booktabs}
\usepackage{hyperref}

\usepackage{amsmath}
\usepackage{amssymb}
\usepackage{mathtools}
\usepackage{amsthm}
\usepackage{tabularx}

\usepackage{amsmath,amsfonts,bm}

\def\eqref#1{equation~\ref{#1}}

\def\1{\bm{1}}

\DeclareMathAlphabet{\mathsfit}{\encodingdefault}{\sfdefault}{m}{sl}
\SetMathAlphabet{\mathsfit}{bold}{\encodingdefault}{\sfdefault}{bx}{n}

\usepackage{url}
\usepackage{multirow}
\usepackage{wrapfig}
\usepackage{pgfplots}
\usepackage{pgfplotstable}
\usepackage{subcaption}
\usepackage{float}
\pgfplotsset{compat=1.18}
\usepgfplotslibrary{groupplots}
\definecolor{IAplotBlue}{HTML}{0072B2}
\definecolor{IAplotOrange}{HTML}{D55E00}
\definecolor{IAplotGray}{HTML}{777777}
\pgfplotsset{
  ia axis/.style={
    scale only axis,
    font=\fontsize{8}{9.5}\selectfont,
    title style={font=\fontsize{9}{11}\selectfont, text=black},
    label style={font=\fontsize{8}{9.5}\selectfont, text=black},
    tick label style={font=\fontsize{8}{9.5}\selectfont, text=black},
    axis lines*=left, axis line style={black!65, line width=.45pt},
    tick align=outside, tick style={black!65, line width=.45pt},
    grid style={black!9, line width=.35pt},
    scaled ticks=false, minor tick num=0,
    legend style={draw=none, fill=none, text=black,
      font=\fontsize{8}{9.5}\selectfont, /tikz/every even column/.append style={column sep=9pt}},
    clip=false,
  },
  ia curve/.style={color=IAplotBlue, solid, line width=1.1pt, mark=*, mark size=1.7pt},
  softmax curve/.style={color=IAplotGray, dashed, line width=.85pt, mark=square*, mark size=1.6pt},
  s4d curve/.style={color=IAplotOrange, dashdotted, line width=.85pt, mark=triangle*, mark size=2pt},
}

\newcommand{\appref}[1]{Appendix~\ref{#1}}
\newcommand{\apptabref}[1]{\mbox{Appendix~\Cref{#1}}}

\usepackage[capitalize,noabbrev]{cleveref}

\newenvironment{talign*}
{\csname align*\endcsname}
{\endalign}

\title{Interdomain Attention:\\Beyond Token-Level Key-Value Memory}

\author{%
Naoki Kiyohara\textsuperscript{*,1,4},
Harrison Bo Hua Zhu\textsuperscript{*,1,2,5},
Riccardo El Hassanin\textsuperscript{*,1}\\
\bfseries Zhuo Sun\textsuperscript{3,1}, Wenlong Chen\textsuperscript{1},
Samir Bhatt\textsuperscript{1,2}, Yingzhen Li\textsuperscript{1}\\
\textsuperscript{1}Imperial College London, UK\\
\textsuperscript{2}University of Copenhagen, Denmark\\
\textsuperscript{3}Shanghai University of Finance and Economics, China\\
\textsuperscript{4}Canon Inc., Japan\\
\textsuperscript{5}Technical University of Denmark, Denmark
}

\iclrfinalcopy

\begin{document}
\maketitle
\lhead{Preprint}

\begingroup
\renewcommand{\thefootnote}{\fnsymbol{footnote}}
\footnotetext[1]{Equal contribution. Correspondence:
Zhuo Sun (\texttt{sunzhuo@mail.shufe.edu.cn}),
Wenlong Chen (\texttt{wenlong.chen21@imperial.ac.uk}),
Yingzhen Li (\texttt{yingzhen.li@imperial.ac.uk}).}
\endgroup

\begin{abstract}
Softmax attention-based transformers and recurrent memory-based models (e.g., state space models (SSMs)) form the foundation of modern sequence models. Hybrid architectures aim to combine the expressivity from attention and the computational efficiency from recurrent memory, yet so far common practices simply stack both attention and recurrent layers. 
We propose Interdomain Attention (IA) as a novel hybrid attention architecture that mixes attention and recurrent memory in the same layer. Motivated by a kernel regression view of softmax attention, IA compresses values and kernel features of keys into a recurrent memory parameterized by an SSM. Then IA approximates dot-product attention by implicitly reconstructing the attention matrix and values from memory states; importantly this construction maintains SSM's computational efficiency advantage. We demonstrate IA's efficacy by designing and pre-training IA network architectures, where ablation studies identify attention-like readout as the primary source of gain over SSMs.
Across 125M-1.3B models pre-trained on text data, IA improves over an S4D baseline at matched memory size. It also outperforms softmax attention-based Transformer in perplexity and accuracy on common-sense reasoning tasks. Based on Caduceus backbone, IA matches or outperforms the original SSM-based architecture in both perplexity and downstream tasks on genomic data.

\end{abstract}

\section{Introduction}

The phenomenal success of GenAI and foundation models is mainly attributed to the invention of scalable sequence modeling architectures for long sequences, e.g., natural language and genomics. In this regard, softmax dot-product attention \citep{vaswani_attention_2017} and recurrent memory such as state space models (SSMs) \citep{gu_hippo_2020,gu_efficiently_2022} are two representative paradigms for sequence model's building blocks. 
Though highly performant, softmax attention-based architectures are limited by their fundamental issue in computational and memory costs: given an input token sequence of length $N$, softmax attention incur $O(N^2)$ computational cost in FLOPs and $O(N)$ memory cost for the key-value (KV) cache required by low-latency inference \citep{shazeer_fast_2019,dao_flashattention_2022,dao_flashattention-2_2024}. 
Methods such as hardware-aware computational kernels~\citep{dao_flashattention_2022, dao_flashattention-2_2024}, speculative decoding~\citep{leviathan_fast_2023} and distributed sharding~\citep{shoeybi_megatron-lm_2020} have been developed to mitigate the issue, but they do not address the fundamental challenge of high computational and memory complexities.

On the other hand, SSMs and Mamba architectures \citep{gu_linear-time_2024, dao_transformers_2024} were proposed as a solution for the quadratic scaling issue. By maintaining a fixed-size recurrent state (which leads to $\mathcal{O}(1)$ memory cost), SSMs update their recurrent memory in $\mathcal{O}(1)$ time per token processing step, thereby reducing the computational cost to $\mathcal{O}(N)$. However, it pays the price of inferior performance to softmax attention architectures in tasks that require precise information recall from the input token sequences.
This performance gap is largely addressed by hybrid architectures \citep{lenz_jamba_2025, glorioso_zamba_2024, ren_samba_2025, de_griffin_2024, fu_hungry_2023} that combine both paradigms. In particular, recent frontier models, such as Evo2~\citep{brixi_genome_2026} for genomics or Qwen3.8-Next~\citep{qiu_design_2026} and Kimi-K3~\citep{team_kimi_2026} for language, use building blocks that stack recurrent memory layers followed by softmax attention layers. Such practice makes it difficult to attribute the importance of each type of layer within the overall architecture, and it often requires extensive architecture search to find the best stacking strategy.

In this paper, we propose Interdomain Attention (IA) as a novel hybrid attention architecture that mixes recurrent memory with dot-product attention in one layer. By interpreting softmax attention as computing predictions from a kernel regression model trained on key-value pairs ~\citep{tsai_transformer_2019, katharopoulos_transformers_2020}, IA represents this regression model's kernel with finite feature maps and then compresses key features and values in evolving recurrent memory states. Key features and values of the historical tokens are then reconstructed \emph{implicitly} for dot-product attention; this implicit attention step maintains the computational efficiency of recurrent memory layers ($\mathcal{O}(1)$ computational cost per token) while leveraging the expressivity of dot-product attention operations. 

Our methodological contributions include (1) the derivation of the IA framework (\Cref{sec:feature-map}) and (2) its practical architecture design (\Cref{sec:architecture}). The headline results from the IA architecture include:
\begin{itemize}
\item On a 125M-1.3B language-modeling study on FineWeb-Edu at matched state size (\Cref{sec:exp-new-lm}), IA achieves better validation perplexity over softmax attention and an S4D baseline~\citep{gu_parameterization_2022} at every scale. At 1.3B IA surpasses softmax attention on seven out of eight commonsense reasoning tasks and on their average \citep{yang_gated_2025}. At 1.3B, IA also achieves higher mean log-likelihood than S4D across 14 recall tasks.
\item On a genomic modeling study (\Cref{sec:exp-genomics}), by replacing the SSM component in the Caduceus backbone \citep{schiff_caduceus_2024} with IA, Caduceus-IA closely matches native Caduceus on the HG38 task, and its parameter-sharing (PS) variant %
Caduceus-IA-PS, which realizes reverse-complement (RC) equivariance, outperforms Caduceus-PS on 14 of 18 Nucleotide Transformer \citep{dalla-torre_nucleotide_2025} downstream tasks. %
\end{itemize}

Ablation studies in both language and genomics tasks show that the attention-like, query-dependent readout of IA blocks is key to the improved performance when compared with SSM-based architectures. Complexity and inference run-time figures also show advantages of IA over softmax attention. All these evidences support IA as a promising hybrid architecture for future sequence models.

\section{Background}\label{sec:background}

Our architecture bridges kernel attention with fixed-size context compression in state space models. We review both frameworks below.

\paragraph{Attention as kernel regression} Standard dot-product attention~\citep{vaswani_attention_2017} projects input tokens $x_n \in \mathbb{R}^d$ into queries, keys, and values $q_n, k_n, v_n \in \mathbb{R}^d$, giving output for token $i$ 
\begin{align}\label{eq:nw}
o_i = \frac{\sum_{n=1}^{N} \mathcal{K}(q_i, k_n)\, v_n}{\sum_{n'=1}^{N} \mathcal{K}(q_i, k_{n'})},
\end{align}
where $\mathcal{K}(q, k) = \exp(q^\top k / \sqrt{d})$ denotes the softmax kernel.
As noted in prior work~\citep{tsai_transformer_2019, katharopoulos_transformers_2020, choromanski_rethinking_2021, wang_test-time_2026}, this formulation corresponds to the Nadaraya--Watson kernel regression estimator~\citep{nadaraya_estimating_1964, watson_smooth_1964}, where attention performs kernel regression over the key-value pairs $\{(k_n, v_n)\}_{n=1}^N$ with the kernel $\mathcal{K}$ and returns the kernel regression model's prediction given query input $q_i$ for $i = 1, ..., N_q$.

\paragraph{State space models and HiPPO} A linear continuous-time state space model (SSM) maps an input signal $z(t) \in \mathbb{R}$ to a latent state $u(t) \in \mathbb{R}^M$ via
\begin{align}\label{eq:ssm}
\dot{u}(t) = A(t)\, u(t) + B(t)\, z(t).
\end{align}
The High-order Polynomial Projection Operators (HiPPO) framework~\citep{gu_hippo_2020} derives transition matrices $A(t) \in \mathbb{R}^{M \times M}$ and $B(t) \in \mathbb{R}^{M}$ that continuously maintain the projection coefficients of the input history onto $M$ time-varying orthogonal polynomial bases $\{\phi_m^{(t)}\}_{m=1}^{M}$. This recurrence allows models to maintain a compressed representation of the context online, forming the foundation of modern deep SSMs~\citep{gu_efficiently_2022, gu_parameterization_2022, gu_linear-time_2024}.

\section{Interdomain Attention}

\definecolor{qblue}{RGB}{30,50,200}
\definecolor{korange}{RGB}{200,110,0}
\definecolor{vviolet}{RGB}{150,40,170}
\definecolor{sgreen}{RGB}{30,140,30}

\begin{figure*}[t]
\centering
\begin{tikzpicture}[
    >=latex,
    tensor/.style={
        draw, rounded corners=2pt,
        minimum height=5mm, minimum width=8mm,
        align=center, inner sep=1.5pt,
        fill=#1, font=\scriptsize
    },
    tensor/.default={gray!12},
    op/.style={
        draw=gray!60, rounded corners=2pt,
        minimum height=6mm, inner sep=3pt,
        align=center, font=\scriptsize,
        fill=white
    },
    mem/.style={
        draw, rounded corners=3pt,
        inner sep=4pt, align=center
    },
    flow/.style={->, thick, draw=black!50},
    qflow/.style={->, thick, draw=qblue},
    kflow/.style={->, thick, draw=korange},
    vflow/.style={->, thick, draw=vviolet},
    sflow/.style={->, thick, draw=sgreen},
    cplx/.style={font=\scriptsize\itshape},
    annot/.style={font=\tiny, text=black!40},
]

\begin{scope}
    \node[font=\small\bfseries] at (0, 4.6) {(a) Standard attention};

    \node[tensor={blue!15}]   (q1) at (-1.3, 3.9) {$Q$};
    \node[tensor={orange!15}] (k1) at ( 0,   3.9) {$K$};
    \node[tensor={violet!15}] (v1) at ( 1.3, 3.9) {$V$};

    \node[mem, fill=red!10,
          minimum height=16mm, minimum width=32mm] (scores) at (-0.65, 1.8) {};
    \node[font=\scriptsize] at (-0.65, 2.05) {%
        $\textcolor{red!60!black}{A} = \operatorname{softmax}\!\!\left(\dfrac{\textcolor{qblue}{Q}\textcolor{korange}{K}^\top}{\sqrt{d}}\right)$};
    \node[cplx, text=red!50!black] at (-0.65, 1.3) {$N_q \times N$ scores};

    \draw[qflow] (q1.south) to[out=-90, in=90] ([xshift=-5mm]scores.north);
    \draw[kflow] (k1.south) to[out=-90, in=90] ([xshift=5mm]scores.north);

    \node[op, fill=gray!6, minimum width=28mm, minimum height=16mm] (mul1) at (0, -0.45) {};
    \node[align=center, font=\scriptsize] at (0, -0.45) {%
        value aggregation\\[2pt]
        ${\color{gray}\underbrace{\textcolor{red!60!black}{A}}_{N_q\times N}} \cdot {\color{gray}\underbrace{\textcolor{vviolet}{V}}_{N\times d}}$};

    \draw[flow] (scores.south) -- ([xshift=-6.5mm]mul1.north);
    \coordinate (vmid) at (1.3, 1.1);
    \draw[vflow, ->] (v1.south) -- (vmid) to[out=-90, in=90] ([xshift=6.5mm]mul1.north);

    \node[tensor] (o1) at (0, -1.8) {Output};
    \draw[flow] (mul1.south) -- (o1.north);
\end{scope}

\begin{scope}[xshift=58mm]
    \node[font=\small\bfseries] at (0.2, 4.6) {(b) Interdomain Attention};

    \node[tensor={blue!15}]   (q2) at (-1.3, 3.9) {$Q$};
    \node[tensor={orange!15}] (k2) at ( 0.0, 3.9) {$K$};
    \node[tensor={violet!15}] (v2) at ( 1.3, 3.9) {$V$};

    \node[op, fill=blue!5] (rq) at (-1.3, 3.1) {$\xi$};
    \node[op, fill=orange!6] (rk) at ( 0.0, 3.1) {$\xi$};
    \draw[qflow] (q2) -- (rq);
    \draw[kflow] (k2) -- (rk);

    \node[op, fill=green!6, minimum width=22mm, minimum height=7mm, align=center] (ssm) at (1.05, 2.1) {%
        SSM\\[-1pt]recurrence};

    \coordinate (ssm_in_l) at ([xshift=-7mm]ssm.north);
    \coordinate (ssm_in_r) at ([xshift=7mm]ssm.north);
    \draw[kflow] (rk.south) to[out=-80, in=90, looseness=1.0] (ssm_in_l);
    \draw[vflow] (v2.south) to[out=-100, in=90, looseness=1.0] (ssm_in_r);

    \node[tensor={blue!8}, font=\scriptsize,
          minimum width=18mm, minimum height=7mm, align=center] (fq) at (-1.3, 1.15) {%
        kernel query\\[-1pt]$\textcolor{qblue}{F_q}$};
    \draw[qflow] (rq) -- (fq);

    \node[mem, fill=green!15,
          minimum height=7mm, minimum width=22mm, inner sep=3pt,
          font=\scriptsize, align=center] (basis) at (1.05, 1.15) {%
          interdomain states\\[-1pt]
          $(\textcolor{korange}{U},\;\textcolor{vviolet}{\Gamma})$};
    \draw[sflow] (ssm.south) -- (basis.north);

    \node[mem, fill=green!4, draw=gray!50,
          minimum width=34mm, minimum height=16mm,
          inner sep=3pt] (ctr) at (-0.1, -0.45) {};
    \node[align=center, font=\scriptsize] at (-0.1, -0.45) {%
        state-space readout\\[2pt]
        ${\color{gray}\underbrace{\textcolor{black}{%
            \textcolor{qblue}{F_q}\;\textcolor{korange}{U}^\top%
        }}_{N_q\times M}}\;\cdot\;{\color{gray}\underbrace{\textcolor{vviolet}{\Gamma}}_{M\times d}}$};

    \coordinate (ctr_in_l) at ([xshift=-10mm]ctr.north);
    \coordinate (ctr_in_r) at ([xshift=10mm]ctr.north);
    \draw[qflow] (fq.south) to[out=-90, in=90, looseness=1.4] (ctr_in_l);
    \draw[sflow] (basis.south) to[out=-90, in=90, looseness=1.4] (ctr_in_r);

    \node[tensor] (o2) at (-0.1, -1.8) {Output};
    \draw[flow] (ctr.south) -- (o2.north);
\end{scope}

\end{tikzpicture}
\caption{%
    (a)~Standard attention computes $N_q \times N$ scores from $Q$ and $K$,
    then multiplies by $V$ to produce the output.
    (b)~Interdomain Attention maps queries and keys into a shared feature space
    to produce the kernel query matrix~$\textcolor{qblue}{F_q}$ ($N_q \times R$).
    Keys and values are compressed via SSM recurrence into $M$~interdomain states:
    $\textcolor{korange}{U}$ ($M \times R$, key feature projections)
    and $\textcolor{vviolet}{\Gamma}$ ($M \times d$, value projections).
    State-space readout computes the output via the $N_q \times M$ interdomain attention matrix
    $\textcolor{qblue}{F_q}\textcolor{korange}{U}^\top$.%
}
\label{fig:architecture}
\end{figure*}

We propose Interdomain Attention (IA) as a new design choice for attention mechanisms from kernel regression perspective.
The IA architecture is inspired by HiPPO-SVGP~\citep{chen_recurrent_2025} as an instance of \emph{interdomain} Gaussian process~\citep{lazaro-gredilla_inter-domain_2009}, where memory of historical datapoints is constructed via functional projections of kernel features (which is referred to as ``interdomain'' kernel features). Therefore, instead of using an explicit KV cache, IA conducts attention with recurrent memory, by projecting the key features and values onto a function space formed by HiPPO/SSM basis functions~\citep{gu_hippo_2020} and computing attention therein. Figure~\ref{fig:architecture} provides an overview of the IA framework, which is motivated by derivations in \Cref{sec:feature-map}. The detailed design, including optional enhancements, are presented in Sections \ref{sec:architecture} and \ref{sec:adaptive}, respectively.

\subsection{Implicit dot-product attention via interdomain kernel features}\label{sec:feature-map}
We first approximate the attention kernel with a finite feature representation $\xi(\cdot)\in\mathbb{R}^{R}$ to give $\mathcal{K}(q,k) \approx \xi(q)^\top \xi(k)$. Substituting this into~\Cref{eq:nw} gives
\begin{align}\label{eq:nw-feature}
o_i\approx \hat{o}_i = \frac{\sum_{n=1}^{N} \xi(q_i)^\top \xi(k_n)\, v_n}{\sum_{n'=1}^{N} \xi(q_i)^\top \xi(k_{n'})}.
\end{align}
This formulation suggests the memory of keys should be constructed for their kernel features, motivating the following design of interdomain kernel features obtained by function projections. 
Specifically, we treat the keys and values as continuous functions of time, $k(t_n) := k_n$ and $v(t_n) := v_n$, and project the key features and the values onto the HiPPO basis $\{\phi_m^{(t_N)}\}_{m=1}^{M}$:
\begin{equation}\label{eq:projection}
u_m^{(t_N)} = \int \xi\!\big(k(t)\big)\, \phi_m^{(t_N)}(t)\, dt,\quad \gamma_m^{(t_N)} = \int v(t)\, \phi_m^{(t_N)}(t)\, dt, \quad \eta_m^{(t_N)} = \int \phi_m^{(t_N)}(t)\, dt,
\end{equation}
where $u_m^{(t_N)} \in \mathbb{R}^{R}$ and $\gamma_m^{(t_N)} \in \mathbb{R}^{d}$. Here $u_m^{(t_N)}$ is the interdomain kernel feature for the $m$-th temporal basis function after observing the $N$-th token \citep{chen_recurrent_2025}, and we also refer to it as key coefficient. 
All three coefficients can be updated online with the HiPPO ODE~\Cref{eq:ssm}.

The key coefficients can then be used to reconstruct the key features as $\xi(k_n) \approx \sum_{m} u_m\, \phi_m(t_n)$.
Substituting this reconstructed $\xi(k_n)$ into~\Cref{eq:nw-feature} and swapping the sums over $m$ and $n$ leaves the token sums in the numerator as $\sum_n \sum_m u_m \phi_m(t_n)\, v_n = \sum_m u_m \sum_n \phi_m(t_n)\, v_n \approx \sum_m u_m \gamma_m$, where the third approximation comes from the HiPPO projection. Similarly for the denominator $\sum_{n} \sum_m u_m \phi_m(t_n) = \sum_{m} u_m \sum_n \phi_m(t_n) \approx \sum_{m} u_m \eta_m$. Dropping the superscript ${}^{(t_N)}$:
\begin{align}\label{eq:interdomain}
\hat{o}_i \approx \tilde{o}_i = \frac{\sum_{m=1}^{M} \big(\xi(q_i)^\top u_m\big)\, \gamma_m}{\sum_{m=1}^{M} \big(\xi(q_i)^\top u_m\big)\, \eta_m}\quad \Rightarrow \quad \tilde{O} = (F_q\, U^\top \Gamma) \oslash (F_q\, U^\top \eta),
\end{align}
with the query features $F_q = [\xi(q_1), ..., \xi(q_{N_q})]^{\top} \in \mathbb{R}^{N_q \times R}$, key coefficients $U = [u_1, ..., u_M]^{\top} \in \mathbb{R}^{M \times R}$, value coefficients $\Gamma = [\gamma_1, ..., \gamma_M]^{\top} \in \mathbb{R}^{M \times d}$ and $\eta = [\eta_1, ..., \eta_M]^\top \in \mathbb{R}^{M}$. Note that the sizes of the recurrent memory states $U$, $\Gamma$ and $\eta$ do not depend on $N$. In causal processing they become position-dependent: $U^{(i)}, \Gamma^{(i)}, \eta^{(i)}$ summarize tokens $1, \ldots, i$.

Compared with softmax attention, IA's key feature reconstruction is conducted \emph{implicitly}, and IA's attention is conducted in a different domain. Here the query features $F_q$ attend to the \emph{interdomain} key kernel features $U$ (instead of kernel features $[\xi(k_1), ..., \xi(k_N)]^\top$ as in softmax attention), which forms the ``interdomain'' attention matrix $F_q U^\top$. ``Retrieval'' is then performed on the function projections $\Gamma$ of the original values. For practical design, we follow linear attention practices \citep{yang_parallelizing_2024} which drop the denominator in softmax attention and compute output as $F_q U^\top \Gamma$ only. A comparison between softmax attention and IA is visualized in Figure~\ref{fig:architecture}. IA is also different from SSM-based architecture due to the construction of recurrent memory that supports an attention-like readout. 

\subsection{A practical implementation of Interdomain Attention}\label{sec:architecture}
We present a practical IA architecture design that instantiates the IA framework described above. The required design choices include kernel feature map, recurrent memory modules for function projections, and other considerations for numerical stability in training and inference.

\paragraph{Kernel feature map}
We use $\xi(\cdot)= \operatorname{L2Norm}\Big( \operatorname{SiLU}\big(\mathrm{SC}(\cdot)\big)\Big)$ proposed by \citet{yang_parallelizing_2024}, where $\mathrm{SC}(\cdot)$ is a causal depthwise 1D convolution applied to queries and keys before the head split, the short convolution used in recent subquadratic models~\citep{fu_hungry_2023, poli_hyena_2023, gu_linear-time_2024}, and $\operatorname{L2Norm}$ is the L2 normalization. $\xi(q)^{\top}\xi(k)$ is then a learned kernel, and we found this design numerically stable at large scale, corroborating reported results in \citet{yang_parallelizing_2024}.

\paragraph{Normalization}
To prevent state explosion and numerical instability in recurrent memory~\citep{yang_parallelizing_2024}, we apply $\operatorname{RMSNorm}$ with learnable biases to the key features and the values:
\begin{equation}\label{eq:inputnorm}
    \tilde{k}_h^{(t)} = \operatorname{RMSNorm}\big(\xi(k_t^{(h)})\big) + b_h^{k},\quad \tilde{v}_h^{(t)} = \operatorname{RMSNorm}(v_t^{(h)}) + b_h^{v}.
\end{equation}
This forms the combined input $z_t^{(h)} = [\tilde{k}_h^{(t)}, \tilde{v}_h^{(t)}]$ to the recurrent memory discussed below. 

\paragraph{Function projection using SSMs}
Per head $h$, we approximate the projections in~\Cref{eq:projection} with a complex diagonal S4D recurrence~\citep{gu_parameterization_2022} using $z_t^{(h)}\in\mathbb{R}^{R+d}$ after normalization:
\begin{align}
s_t^{(h)} = \mathrm{diag}(\Lambda_h) s_{t-1}^{(h)} + B_h\, (z_t^{(h)})^\top,\quad \Lambda_h =& \exp(\Delta_h A_h) \in \mathbb{C}^{M}, \ B_h=(\exp(\Delta_h A_h)-1) \oslash A_h, \nonumber \\
 [U_h^{(t)},\,\Gamma_h^{(t)}] =& 2\operatorname{Re}\!\left(C_h s_t^{(h)}\right)
\in\mathbb{R}^{M\times(R+d)}, \quad C_h \in \mathbb{C}^{M\times M}.
\end{align}
Here $A_h$ adopts the diagonal S4D-Inv initialization~\citep{gu_parameterization_2022} approximating HiPPO-LegS dynamics, and $A_h$, $\Delta_h$ and $C_h$ are learned end-to-end.

\paragraph{Implementation}
For efficient training, we adapt chunkwise execution from Flash Linear Attention~\citep{yang_gated_2024}, combining parallel computation within each chunk with sequential state passing across chunks to keep total work linear in sequence length. We develop optimized chunkwise kernels in Triton for language-modeling scaling experiments and in TileLang~\citep{wang_tilelang_2026} for ablations (\appref{app:impl}, \appref{app:adaptive-details}). At inference time, the fixed-size recurrent state supports CUDA-graph capture and chunked prefill, enabling constant-memory token generation (\appref{app:inference-scaling}).

\paragraph{Complexity figures}
In training, IA achieves $\mathcal{O}(N)$ computational complexity in sequence length $N$, while standard softmax attention scales as $\mathcal{O}(N^2)$. Memory complexity for IA is $\mathcal{O}(N)$, as well as softmax attention with FlashAttention.
During inference, softmax attention requires an expanding $\mathcal{O}(Nd)$ KV cache and $\mathcal{O}(Nd)$ computation per token generation. In contrast, IA compresses history tokens into recurrent states of size $\mathcal{O}(M(R{+}d))$ independent of $N$. Each decoding step takes $\mathcal{O}(M(R+d)+M^2)$ work, enabling constant-memory generation and CUDA-graph capture (decode latency and prefill memory benchmarks are reported in \appref{app:inference-scaling}). A detailed comparison of the complexity figures can be found in Table \ref{tab:complexity}.

\begin{table}[t]
\centering
\caption{Per-head computational and memory cost comparisons. FA denotes FlashAttention~\citep{dao_flashattention_2022}. For IA the figures are reported for full $C_h$, and $K$ is the chunk length.}
\label{tab:complexity}
\small
\begin{tabular}{lccc}
\toprule
 & Computational cost & Memory (inference) & Saved activations (training) \\
\midrule
Softmax (FA) & $\mathcal{O}(N^2d)$ & $\mathcal{O}(Nd)$ & $\mathcal{O}(Nd)$ \\
IA (ours) & $\mathcal{O}(N[M(R{+}d)+M^2])$ & $\mathcal{O}(M(R{+}d))$ & $\mathcal{O}(\tfrac{N}{K}M(R{+}d)+N(R{+}d{+}M))$ \\
\bottomrule
\end{tabular}
\end{table}

\subsection{Optional Architectural Enhancements}\label{sec:adaptive}

In addition to the base architecture, we explore optional refinements to enhance model expressivity without increasing the recurrent memory size. They are applied selectively to IA models trained on language and genomic sequence data.

\paragraph{Non-linear query projection}
While linear projection $q_t = W_q i_t$ given layer input $i_t\in\mathbb{R}^{d_{\mathrm{model}}}$ is commonly used, improved query construction can also enhance the expressivity of the network layer. Hence we introduce an optional nonlinear query by adding a residual MLP to the projection:
\begin{equation}\label{eq:adaptive-query}
    q_t = W_q i_t + W_{q,2}\,\operatorname{SiLU}(W_{q,1}i_t).
\end{equation}
We initialize $W_{q,2}$ to zero so that training begins from the standard linear projection.

\paragraph{Value paths and output gating}
We incorporate three refinements established in recent recurrent memory architectures~\citep{gu_parameterization_2022, gu_linear-time_2024, yang_gated_2024}.
First, in trained IA models for genomics data we add a causal depthwise convolution followed by $\operatorname{SiLU}$ activation to construct the values $v_t^{(h)}$ before normalization.
Second, we add a direct skip connection $D_h\odot v_t^{(h)}$ to the readout, initializing $D_h\in\mathbb{R}^{d}$ to ones.
Finally, an output gate $g_t^{\text{out}} = \operatorname{SiLU}(W_g i_t)$ is applied per head before the final output merge across heads.

\paragraph{Selective writing}
Following practices in modern recurrent memory architectures~\citep{gu_linear-time_2024}, we introduce input-dependent gating to control the information for memory storage. We define a positive write gate $g_t^{\mathrm{write}}
=\exp\!\left[\alpha\odot\tanh\!\left(w\odot\operatorname{RMS}_0(\bar{k}_t)+b\right)\right]$, where $\bar{k}_t$ is kernel feature map for key just before $\operatorname{L2Norm}$,
$\operatorname{RMS}_0$ is RMS normalization without learned parameters, and $w,b,\alpha\in\mathbb R^d$ are per-head learned parameters, with $R=d$.
It scales both the key features and values before they enter the recurrence: $z_t^{\mathrm{write}} =\bigl[\sqrt{g_t^{\text{write}}}\odot\tilde{k}_t,\; \sqrt{g_t^{\text{write}}}\odot\tilde{v}_t\bigr]$.
We apply the same square-root gate for both key and value so that they are jointly modulated.

\paragraph{Selective memory decay}
Following Mamba~\citep{gu_linear-time_2024}, we dynamically adjust memory decay based on the input. For each mode with base transition $\Lambda_m = \rho_m \exp(\mathrm{i}\theta_m)$, an input-dependent $\delta_t > 0$ rescales the decay rate for memory modulation:
\begin{equation}\label{eq:selective-temporal-update}
    \Lambda_{t,m} = \rho_m^{\delta_t} \exp(\mathrm{i}\theta_m), \quad s_t = \Lambda_t \odot s_{t-1} + B\,(z_t^{\mathrm{write}})^\top,
\end{equation}
where, omitting the head index,
  $\delta_t=\exp\!\left[(\log 2)\tanh\!\left(
  w_\delta^\top z_t^{\mathrm{write}}+b_\delta
  \right)\right]$,
  with learned per-head parameters $w_\delta$ and $b_\delta$.

\color{black}

\section{Related Work}\label{sec:related-works}

\paragraph{Efficiency in attention mechanisms}
The quadratic cost of attention has motivated both faster implementations and alternatives to avoid explicit pairwise token interactions. At the implementation level, FlashAttention uses tiling to compute exact softmax attention without storing the full attention matrix, reducing GPU memory transfers while retaining quadratic arithmetic complexity \citep{dao_flashattention_2022}. Hardware-efficient Linear Attention (see next paragraph for detailed discussion in linear attention), Flash Linear Attention (FLA), uses chunkwise parallel computation and kernel fusion to improve GPU utilization and reduce memory overhead \citep{yang_gated_2024}. Similarly, our IA also adopts hardware-efficient chunkwise execution. At the algorithmic level, Performer uses positive random features whose inner products approximate the softmax kernel with linear time and space complexity \citep{choromanski_rethinking_2021}.

\paragraph{SSMs and Linear Attention}
SSMs and linear attention summarize history in a recurrent state whose size does not grow with sequence length, with state space duality formalizing connections between the two \citep{dao_transformers_2024}. SSMs describe the evolution of a latent state through structured dynamics: HiPPO compresses sequence history into a fixed-size memory through online basis projections \citep{gu_hippo_2020}, providing the foundation for S4, a structured state-space sequence model, and its diagonal variant S4D \citep{gu_efficiently_2022,gu_parameterization_2022}, while Mamba introduces input-dependent state updates \citep{gu_linear-time_2024}. Linear attention maintains key-value associations for query-dependent retrieval. Naive linear Transformers use finite feature maps to accumulate key--value statistics in a fixed-size state \citep{katharopoulos_transformers_2020}. More expressive linear attentions include Gated Linear Attention which controls memory retention \citep{yang_gated_2024} and Gated DeltaNet which combines gating with delta-rule updates \citep{yang_gated_2025}. Our IA builds on the projection view of memory to represent keys and values through projection coefficients corresponding to a shared temporal basis, and we combine these key and value representations at readout rather than accumulating their products.

\paragraph{Hybrid architectures}
Existing hybrid architectures combine recurrent or convolutional layers with attention mainly by stacking them: Jamba pairs Mamba layers with attention and uses mixture-of-experts modules in selected feedforward layers \citep{lenz_jamba_2025}, while Griffin combines gated linear recurrent units with sliding-window attention to compress history over a bounded cache of recent token representations for direct retrieval \citep{de_griffin_2024}. Recent language models extend this approach: Qwen3.8-Next combines Gated DeltaNet layers with token-level attention, using Qwen Sparse Attention during continued pretraining \citep{qiu_design_2026}, while Kimi K3 combines Kimi Delta Attention with gated Multi-head Latent Attention \citep{team_kimi_2026}. In genomics, Evo~2 uses StripedHyena~2 to combine Hyena convolutional operators with attention layers \citep{brixi_genome_2026}. H3 instead combines SSMs and multiplicative interactions within a layer, compressing key--value products for query-dependent retrieval \citep{fu_hungry_2023}. Our IA is a hybrid attention operator that integrates recurrent compression and content-based retrieval within a single layer, rather than stacking different layers.

\section{Experiments}\label{sec:experiments}

We evaluate the proposed IA architecture by pre-training and fine-tuning models on language and genomic data. The major baselines include SSM and softmax attention-based architectures, where our primary goal is to understand the benefits brought by the hybrid scheme of attention-like readout with recurrent memory. Architecture details can be found in \appref{app:implementation-training}. In the main text we report summaries of the empirical results; detailed results can be found in Appendix.

\subsection{Language modeling}\label{sec:exp-new-lm}

\paragraph{Architecture and data}
We use Llama-3 backbone~\citep{grattafiori_llama_2024} and replace the multi-head attention layer in pre-norm decoder with IA. We train four models with 125M, 350M, 760M, and 1.3B parameters on FineWeb-Edu dataset \citep{penedo_fineweb_2024}, with Chinchilla-optimal token budgets of 2.5B, 7B, 15B, and 26B tokens respectively~\citep{hoffmann_empirical_2022}. We use Llama-2 tokenizer~\citep{touvron_llama_2023-1}, and context length 4096.

\paragraph{Scaling curve}
\Cref{fig:main-lm-scaling-context}(a) shows best-of-run validation perplexity for Softmax, S4D-only, and IA across four model scales on FineWeb-Edu. IA outperforms the matched-state S4D-only baseline at every scale. At the largest 1.3B scale, IA achieves a validation perplexity of $7.98$, outperforming the Softmax baseline ($8.62$) and S4D-only ($9.13$). The precise numerical results and exact parameter counts are presented in \apptabref{tab:lm-scaling} and \appref{app:implementation-training}, respectively.

\begin{figure}[t]
\centering
\begingroup\color{black}
\begin{tikzpicture}
\begin{groupplot}[
  ia axis, width=.385\linewidth, height=1.28in,
  group style={group size=2 by 1, horizontal sep=.15\linewidth},
  ymajorgrids=true,
]
\nextgroupplot[
  title={(a) FineWeb-Edu scaling},
  xmode=log, log basis x=10, xmin=110, xmax=1450,
  xtick={125,350,760,1300}, xticklabels={125M,350M,760M,1.3B},
  ymode=log, log basis y=10, ymin=7.3, ymax=20,
  ytick={8,10,12,16,20}, yticklabels={8,10,12,16,20},
  xlabel={Number of parameters}, ylabel={Validation PPL},
  legend to name=iaScalingLegend, legend columns=3,
]
\addplot[softmax curve] table[x=softmax_params_m,y=softmax] {figures/experiments/scaling.dat};
\addlegendentry{Softmax}
\addplot[ia curve] table[x=ia_params_m,y=ia] {figures/experiments/scaling.dat};
\addlegendentry{Interdomain Attention}
\addplot[s4d curve] table[x=s4d_params_m,y=s4d] {figures/experiments/scaling.dat};
\addlegendentry{S4D-only}
\nextgroupplot[
  title={(b) Length extrapolation},
  xmin=3.5, xmax=14.6, xtick={4,8,14},
  ymin=0, ymax=66, ytick={0,20,40,60},
  xlabel={Context length (K tokens)}, ylabel={Six-corpus mean PPL},
]
\draw[IAplotGray, densely dotted, line width=.7pt] (axis cs:4,0) -- (axis cs:4,66);
\node[anchor=north west, text=IAplotGray, font=\fontsize{7}{8}\selectfont]
  at (axis cs:4.3,65) {4K training context};
\addplot[softmax curve] table[x=x,y=softmax] {figures/experiments/context.dat};
\addplot[s4d curve] table[x=x,y=s4d] {figures/experiments/context.dat};
\addplot[ia curve] table[x=x,y=ia] {figures/experiments/context.dat};
\node[anchor=south east, inner sep=2pt] at (axis cs:14,62.06) {62.06};
\node[anchor=south east, inner sep=2pt] at (axis cs:14,28.34) {28.34};
\node[anchor=south east, inner sep=2pt, text=IAplotBlue] at (axis cs:14,15.71) {15.71};
\end{groupplot}
\path (group c1r1.north west) -- node[above=22pt] {\pgfplotslegendfromname{iaScalingLegend}} (group c2r1.north east);
\end{tikzpicture}
\caption{%
(a)~Validation perplexity across model scales on FineWeb-Edu (exact values in \apptabref{tab:lm-scaling}). IA outperforms S4D-only at all scales and surpasses Softmax at 1.3\,B.
(b)~Context length extrapolation up to 14K tokens at 1.3\,B (per-corpus breakdowns in \apptabref{tab:lm-length-extrap-per-corpus}). IA remains flat with respect to length, whereas Softmax degrades sharply.%
}
\label{fig:main-lm-scaling-context}
\endgroup
\end{figure}

\paragraph{Downstream evaluation}
We evaluate the 1.3\,B models with the EleutherAI Language Model Evaluation Harness~\citep{biderman_lessons_2026} on the standard eight-task commonsense reasoning suite~\citep{yang_gated_2025}
, alongside language-modeling perplexities on LAMBADA and WikiText-2. As shown in \Cref{tab:new-lm-headline}, IA outperforms the Softmax baseline by $+3.03\%$
on the 8-task commonsense average and achieves significantly lower perplexity on both LAMBADA and WikiText-2. The S4D-only baseline underperforms both models. Details are provided in \appref{app:lm-commonsense-breakdown}. On associative recall, IA achieves higher mean answer log-likelihood than S4D-only for all 14 settings (\appref{app:lm-recall-likelihood}).

\begin{table}[t]
\centering\small
\caption{LM downstream evaluation at 1.3\,B. Commonsense is the eight-task average accuracy (percent); full per-task breakdowns are in \apptabref{tab:lm-commonsense-breakdown}.}
\label{tab:new-lm-headline}
\begin{tabular}{lccc}
\toprule
Model & Commonsense $\uparrow$ & LAMBADA PPL $\downarrow$ & WikiText-2 PPL $\downarrow$ \\
\midrule
Softmax & 51.51 & 21.03 & 20.55 \\
S4D-only & 50.23 & 41.02 & 25.27 \\
Interdomain Attention & \textbf{54.54} & \textbf{14.76} & \textbf{19.77} \\
\bottomrule
\end{tabular}
\end{table}

\paragraph{Length extrapolation}
IA avoids the positional-degradation issues inherent to attention with RoPE due to the use of recurrent memory. When evaluated at context lengths up to $14\text{K}$ ($3.5\times$ the $4\text{K}$ training context) across six long-context corpora (\Cref{fig:main-lm-scaling-context}(b)), IA at 1.3\,B maintains nearly constant perplexity, whereas the Softmax baseline degrades sharply from $14.21$ to $62.06$ at $14\text{K}$. Detailed result breakdown and associative-recall evaluations are in Appendices~\ref{app:lm-length-extrap} and~\ref{app:lm-recall}.

\paragraph{Inference speed}
We use CUDA-graph capture to measure IA's autoregressive decoding throughput. PyTorch SDPA with FlashAttention2 backend~\citep{dao_flashattention-2_2024} is used for softmax attention. For short prefix length and small batch size, highly optimized SDPA keeps constant decode latency with respect to the prefix length. However, for long prefix length (e.g.,~8k) and large batch size (e.g.~8), SDPA's latency starts growing while IA's stays constant. At 1.3\,B, the speedup of IA over Softmax is $1.11\times$ at batch size 1 and $0.69$--$0.94\times$ at batch sizes 8 to 64. Detailed speed results are reported in \appref{app:inference-scaling}.

\begin{figure}[!t]
\centering
\begingroup\color{black}
\colorlet{IArankBlue}{IAplotBlue!75!black}
\newcommand{\IArankstyle}[1]{%
  \ifdim #1pt>3pt
    \def\rankshade{0}\def\ranktext{black}%
  \else
    \pgfmathsetmacro{\rankshade}{100-40*(#1-1)}%
    \ifdim #1pt<2pt\def\ranktext{white}\else\def\ranktext{black}\fi
  \fi
}
\newcommand{\IArankcell}[3]{%
  \IArankstyle{#3}%
  \fill[IArankBlue!\rankshade!white, rounded corners=.7pt]
    ({10.5*#1+.5},{-12*#2-.5}) rectangle ({10.5*(#1+1)-.5},{-12*(#2+1)+.5});
  \node[font=\fontsize{6.5}{7}\selectfont, text=\ranktext, inner sep=0pt]
    at ({10.5*(#1+.5)},{-12*(#2+.5)}) {#3};%
}
\newcommand{\IAranktask}[2]{%
  \node[anchor=east, rotate=60, font=\fontsize{6.5}{7.5}\selectfont, inner sep=0pt]
    at ({10.5*(#1+.5)},-99) {#2};%
}
\newcommand{\IArankmodel}[2]{%
  \node[anchor=east, font=\fontsize{7}{8}\selectfont, inner sep=0pt] at (-3,{-12*(#1+.5)}) {#2};%
}
\begin{tikzpicture}[x=1pt,y=1pt]
\node[anchor=west,font=\fontsize{9}{11}\selectfont] at (0,13) {(a) GenomicBenchmarks};
\node[anchor=west,font=\fontsize{9}{11}\selectfont] at (151,13) {(b) Nucleotide Transformer};
\begin{scope}[shift={(55,0)}]
\input{figures/experiments/genomic-ranks-gb}
\IArankmodel{0}{CNN}
\IArankmodel{1}{HyenaDNA}
\IArankmodel{2}{Mamba}
\IArankmodel{3}{Cad. w/o equiv.}
\IArankmodel{4}{Caduceus-PS}
\IArankmodel{5}{Caduceus-Ph}
\IArankmodel{6}{\textcolor{IAplotBlue}{\textbf{IA-PS (ours)}}}
\IArankmodel{7}{\textcolor{IAplotBlue}{\textbf{IA-Ph (ours)}}}
\draw[black!35,line width=.35pt] (-1,-72) -- (85,-72);
\end{scope}
\begin{scope}[shift={(205,0)}]
\input{figures/experiments/genomic-ranks-nt}
\IArankmodel{0}{Enformer}
\IArankmodel{1}{DNABERT-2}
\IArankmodel{2}{NT-v2}
\IArankmodel{3}{HyenaDNA}
\IArankmodel{4}{Caduceus-PS}
\IArankmodel{5}{Caduceus-Ph}
\IArankmodel{6}{\textcolor{IAplotBlue}{\textbf{IA-PS (ours)}}}
\IArankmodel{7}{\textcolor{IAplotBlue}{\textbf{IA-Ph (ours)}}}
\draw[black!35,line width=.35pt] (-1,-72) -- (190,-72);
\end{scope}
\node[anchor=east,font=\fontsize{7}{8}\selectfont] at (146,-152) {Rank (1 = best)};
\foreach \rank in {1,...,8}{%
  \IArankstyle{\rank}%
  \fill[IArankBlue!\rankshade!white] ({151+11*(\rank-1)},-157) rectangle ({161+11*(\rank-1)},-147);
  \node[font=\fontsize{6.5}{7}\selectfont,text=\ranktext,inner sep=0pt] at ({156+11*(\rank-1)},-152) {\rank};
}
\end{tikzpicture}
\caption{%
Task-wise model rankings on (a)~GenomicBenchmarks and (b)~Nucleotide Transformer (1 = best rank; top-3 shaded). Caduceus-IA variants rank among the top models across most tasks. Full scores and metrics are in \apptabref{tab:gb,tab:nt}.%
}
\label{fig:genomic-downstream-ranks}
\endgroup
\end{figure}

\subsection{Genomic sequence modeling}\label{sec:exp-genomics}

\paragraph{Architecture}
We use Caduceus backbone ~\citep{schiff_caduceus_2024} and replace its Mamba layers with IA, preserving its bidirectional processing and reverse-complement handling. We refer to \emph{Caduceus-IA-PS} as the IA version of Caduceus that uses Parameter Sharing (PS) to preserve reverse-complement (RC) equivariance, which corresponds to Caduceus-PS, and \emph{Caduceus-IA-Ph} as post-hoc (Ph) RC handling and data augmentation, which corresponds to Caduceus-Ph.%

\paragraph{Pre-training} 
We pre-train Caduceus-IA on the human reference genome (HG38) ~\citep{genome_reference_consortium_genome_2009} using bidirectional masked language modeling with the standard BERT masking recipe \citep{devlin_bert_2019}. After 10,000 training steps at a context length of 1024, Caduceus-IA reaches a validation perplexity of 2.614, closely matching native Caduceus at 2.612 under the same context length and training budget (\apptabref{tab:dna-pretraining}; see \appref{app:native-dna-settings} for baseline settings). We also pretrain a separate 7.7M parameters Caduceus-IA configuration with a context length of 131,072 tokens for 50,000 updates.

\paragraph{Downstream evaluation.}
We evaluate Caduceus-IA-PS and Caduceus-IA-Ph against published baselines~\citep{schiff_caduceus_2024} on 8 GenomicBenchmarks regulatory element classification tasks~\citep{gresova_genomic_2023}, 18 Nucleotide Transformer tasks such as histone marker prediction and splice site annotation ~\citep{dalla-torre_nucleotide_2025}, and causal-eQTL variant-effect prediction ~\citep{avsec_effective_2021,kao_advancing_2024}, which assesses whether SNPs affect gene expression. To match the Caduceus experiments, we use Caduceus-IA models pretrained at 1K context for GenomicBenchmarks (model dimension 118, 470K parameters) and Nucleotide Transformer (model dimension 256, 1.94M parameters), and at 131K context for variant-effect prediction (model dimension 256, 7.7M parameters). Here, model dimension refers to the dimension of each token's embedding.

\Cref{fig:genomic-downstream-ranks} summarizes the classification results through task-wise ranks, where detailed numerical results are provided in Appendix Tables \ref{tab:gb} and \ref{tab:nt}. On \textbf{GenomicBenchmarks}, Caduceus-IA-Ph achieves the highest reported mean accuracy on three of eight tasks and shares the lead on two more, while published Caduceus baselines perform best on the remaining three. On \textbf{Nucleotide Transformer}, Caduceus-IA-PS and Caduceus-IA-Ph outperform their corresponding Caduceus baselines on 14 and 12 of 18 tasks, respectively. Caduceus-IA-Ph leads the full comparison on eight H3 and H4 tasks, while NT-v2 retains the lead in all promoter and splice-site tasks. 

For \textbf{variant-effect prediction}, we group variants by distance to the nearest transcription start site. Caduceus-IA-PS outperforms HyenaDNA and NT-v2 in reported mean AUROC across all groups. Against their respective baselines, Caduceus-IA-PS matches performance at 0.679 within 30\,kb, achieves a higher AUROC of 0.620 versus 0.612 at 30--100\,kb, and a lower 0.574 versus 0.586 beyond 100\,kb, while Caduceus-IA-Ph underperforms Caduceus-Ph across all groups (\apptabref{tab:vep}).

\subsection{Analysis of the Interdomain Attention architecture}\label{sec:exp-mixer-analysis}

\paragraph{Mechanism ablation}
To isolate the contributions of the base IA components, we separately ablate two core design choices.
(1) For input structure, we compare how we make the feature $z_t$. In our base IA, before normalization of \Cref{eq:inputnorm}, we use key feature $\xi (k_t)$ and value $v_t$.
The baseline instead uses linear projections $a_t = W_a i_t$ and $b_t = W_b i_t$ of the layer input, without the key feature map. %
(2) For readout, we compare our attention-like, query-conditioned readout with linear readout (see \Cref{fig:mechanism-arch} in \appref{app:mechanism-arch}). In ablations we use the base IA configuration (details in \appref{app:mechanism-inputs}). \Cref{tab:current-mechanism} shows that the attention-like, query-conditioned readout is the largest contributor to the improved performance.
This justifies our core design of attention-like readout motivated by the IA framework which implicitly constructs approximations to dot-product attention.

\begin{table}[t]
\centering\small
\setlength{\tabcolsep}{4pt}
\caption{Mechanism ablation with common initialization and
final-checkpoint evaluation: validation NLL ($\downarrow$).
Non-selective/Selective denotes memory updates with selective
writing and selective memory decay jointly disabled/enabled.
Input preprocessing is described in \appref{app:mechanism-inputs};
additional studies with different evaluation settings are reported
in Appendices~\ref{app:dna-mechanism-minimal}
and~\ref{app:mechanism-decomp}.}
\label{tab:current-mechanism}
\begin{tabular}{@{}llcccc@{}}
\toprule
 & & \multicolumn{2}{c}{LM}
   & \multicolumn{2}{c}{Caduceus} \\
Input & Readout
 & Non-selective & Selective
 & Non-selective & Selective \\
\midrule
Key feature/value
 & Query-conditioned & 2.7949 & 2.7867 & 0.9893 & 0.9892 \\
Key feature/value
 & Linear            & 3.0073 & 3.0002 & 1.0003 & 0.9939 \\
Linear projection
 & Query-conditioned & 2.8049 & 2.7964 & 0.9786 & 0.9783 \\
Linear projection
 & Linear            & 2.9902 & 2.9820 & 0.9802 & 0.9788 \\
\bottomrule
\end{tabular}
\end{table}

\paragraph{Adding optional enhancements}
We next evaluate how each architectural enhancement methods introduced in \Cref{sec:adaptive} contributes to performance.
Starting from a base IA model that omits short convolutions, we incorporate short convolutions to evaluate the effect of short convolution in the base IA,
and added a residual query MLP, a direct skip connection for value,
an output gate, selective writing, and selective memory decay in an incremental manner.
In genomics experiments we additionally include
depthwise convolution followed by SiLU, as mentioned in Section~\ref{sec:adaptive}.
\Cref{fig:main-incremental-ablation} shows that for language modelling most of the architectural enhancements, apart from the value skip connection and selective writing, provide decent improvements for the performance. For genomics, Q/K conv, value conv + SiLU activation and residual query MLP provide most of the gains. The final configurations reduce validation perplexity from 16.595 to 16.008 in language modeling and from 2.737 to 2.614 in genomic modeling.

\begin{figure}[!t]
\centering
\begingroup\color{black}
\begin{tikzpicture}
\begin{groupplot}[
  ia axis, width=.305\linewidth, height=1.6in,
  group style={group size=2 by 1, horizontal sep=.07\linewidth},
  xmin=-.15, xmax=5.8, xtick={0,1,2,3,4,5}, xmajorgrids=true,
  ymin=-.4, ymax=7.45, y dir=reverse,
  ytick={0,1,2,3,4,5,6,7},
]
\nextgroupplot[
  title={(a) Language modeling},
  yticklabels={Base IA without Q/K conv.,Base IA (with Q/K conv.),+ Value conv.\ + SiLU (genomics only),+ Residual query MLP,+ Value skip connection,+ Output gate,+ Selective writing,+ Selective memory decay},
  yticklabel style={font=\fontsize{8}{9.5}\selectfont, text=black},
]
\addplot[ia curve] table[x=gain,y=stage] {figures/experiments/ladder-lm.dat};
\node[anchor=west, text=IAplotBlue, inner sep=3pt] at (axis cs:3.52632,7) {3.53\%};
\nextgroupplot[title={(b) Genomic modeling}, yticklabels=\empty]
\addplot[ia curve] table[x=gain,y=stage] {figures/experiments/ladder-dna.dat};
\node[anchor=west, text=IAplotBlue, inner sep=3pt] at (axis cs:4.51490271,7) {4.51\%};
\end{groupplot}
\path (group c1r1.south west) -- node[below=18pt, font=\fontsize{8}{9.5}\selectfont] {Cumulative PPL reduction (\%)} (group c2r1.south east);
\end{tikzpicture}
\caption{%
Cumulative perplexity reduction from sequentially adding architectural refinements in (a)~language modeling and (b)~genomic sequence modeling. (Details in \apptabref{tab:adaptive-ladder})
}
\label{fig:main-incremental-ablation}
\endgroup
\end{figure}

\section{Conclusion and Future Work}

We propose Interdomain Attention (IA) that approximates dot-product attention by function projection of key features and values into a recurrent memory. At matched recurrent state size, base Interdomain Attention improves over an S4D baseline from 125M to 1.3B parameters; at 1.3B, it also outperforms softmax attention on validation perplexity and downstream common-sense reasoning tasks. Using the Caduceus backbone, IA-based architectures also show competitive or improved performance. Ablation studies across language and genomics identify the attention-like readout as the primary source of performance gains.
For limitations, IA's training throughput is 2.8-times slower than that of PyTorch Flash SDPA, primarily due to less well-optimized GPU kernel when compared with FlashAttention2 (\appref{app:training-speed}). Future work includes building interdomain projections on more recent recurrent memories such as Mamba-3 and Gated DeltaNet instead of S4D,  GPU kernel optimization, and formalizing a unified probabilistic framework to the connection between kernel attention and interdomain Gaussian processes~\citep{chen_calibrating_2023,chen_recurrent_2025,chen_probabilistic_2026}.

\section*{Reproducibility Statement}
All details regarding algorithms and experiments are enumerated in this manuscript. In the event that the paper is accepted, we will open source model weights and code. 

\section*{Ethics Statement}
This work proposes a general-purpose sequence model architecture and evaluates
it on text and genomic modeling tasks. It involves no human subjects research
and no new data collection. All datasets are publicly available and used in
accordance with their licenses. As with other foundation models for biological sequences, improved genomic
sequence modeling is dual-use: it may advance the study of gene regulation
and variant effects, but could in principle contribute to the design of
harmful biological sequences. 

\section*{AI Use Statement}
The authors developed the core idea and mathematical derivations of Interdomain Attention without AI assistance. The authors used Codex, Claude Code, and Gemini to propose and test stabilization refinements and incremental ablations, suggest training settings, and assist with implementing Interdomain Attention, including GPU kernels and experimental pipelines. The tools also helped check the author-written main text and experiments for accuracy, polish the main-text wording, draft Appendix explanations from author outlines, create figures and tables, and identify potentially missing references. The authors checked GPU kernel outputs and gradients against a reference PyTorch implementation and verified AI-assisted text, figures, tables, and reported results against the implementation and experiment records. The authors take responsibility for the final content of this paper.
\bibliography{references}
\bibliographystyle{iclr2027_conference}

\newpage
\appendix

\section{Implementation and training details}\label{app:implementation-training}

\subsection{Architecture details}\label{app:arch-details}
For language modeling, we adopt a Llama-style pre-norm decoder backbone~\citep{touvron_llama_2023}, implemented by adapting the Llama~3 reference code~\citep{grattafiori_llama_2024}, with RMSNorm~\citep{zhang_root_2019}, SwiGLU feedforward layers~\citep{shazeer_glu_2020}, rotary position embeddings (RoPE)~\citep{su_roformer_2024} on the query/key inputs, untied embeddings, and no dropout. For genomic modeling, we use the Caduceus backbone~\citep{schiff_caduceus_2024} with bidirectional weight tying and replace its Mamba layers with IA. Caduceus-IA-PS adds reverse-complement parameter sharing, and Caduceus-IA-Ph uses post-hoc reverse-complement handling with data augmentation (\Cref{sec:exp-genomics}). The HG38 ablations use the PS configuration.
Scale-specific LM dimensions are given in \appref{app:lm-training-backbone}, and the ablation configurations in \appref{app:adaptive-details}.

The short convolution $\mathrm{SC}$ in \Cref{sec:architecture} has kernel size 4 in the language models and 8 in the genomic models; the genomic value convolution also has kernel size 8. $A_h$ uses the S4D-Inv initialization~\citep{gu_hippo_2020,gu_parameterization_2022}, $\operatorname{Re}(A_h) = -\tfrac{1}{2}$ and $\operatorname{Im}(A_h)(n) = \tfrac{M}{\pi}\big(\tfrac{M}{2n+1}-1\big)$, and $\Delta_h$ is initialized log-uniformly in $[10^{-3}, 10^{-1}]$. The readout $C_h$ is a full complex $M\times M$ matrix per head. The selective decay factor $\delta_t$ in \Cref{eq:selective-temporal-update} lies in $[0.5, 2]$ and rescales only the modulus $\rho_m$ of each mode, while the phase $\theta_m$ is unchanged.

\subsection{SSM kernels}\label{app:impl}

IA models are trained with chunkwise kernels for the S4D recurrence~\citep{gu_efficiently_2022,gu_parameterization_2022} that follow Flash Linear Attention~\citep{yang_gated_2024}. The sequence is split into chunks of length $K$. Each chunk computes its final state in parallel, the states are passed serially across the $N/K$ chunk boundaries, and a second parallel pass recomputes each chunk from its corrected initial state. The language-modeling scaling runs use Triton kernels, and the ablations (\appref{app:adaptive-details}) and genomic models use TileLang 0.1.14 kernels~\citep{wang_tilelang_2026}. The chunk length is $K=64$ unless stated otherwise.

\subsection{Language modeling training}\label{app:lm-training-details}

\paragraph{Backbone details.}\label{app:lm-training-backbone}
SwiGLU hidden dimensions are set to $\tfrac{2}{3}\cdot 4d$ rounded up to a multiple of $128$. Residual-path output projections ($w_o$ of attention and $w_2$ of SwiGLU) are re-initialized with standard deviation $0.02/\sqrt{2n_{\mathrm{layer}}}$ for $n_{\mathrm{layer}}$ layers, following the GPT-style residual scaling rule~\citep{radford_language_2019}. \Cref{tab:lm-dims} lists the per-scale dimensions. Softmax, S4D-only and IA share these dimensions at each scale; all use context length $4096$, S4D state dimension $M{=}64$, and feature dimension $R{=}d_h$.

\begin{table}[!htbp]
\centering
\caption{Language-model dimensions per scale: model width $d$, number of layers $n_{\mathrm{layer}}$, heads $H$, head dimension $d_h$, and S4D state dimension $M$.}
\label{tab:lm-dims}
\small
\begin{tabular}{l c c c c c}
\toprule
Scale & $d$ & $n_{\mathrm{layer}}$ & $H$ & $d_h$ & $M$ \\
\midrule
125\,M & $768$ & $12$ & $12$ & $64$ & $64$ \\
350\,M & $1024$ & $24$ & $16$ & $64$ & $64$ \\
760\,M & $1536$ & $24$ & $16$ & $96$ & $64$ \\
1.3\,B & $2048$ & $24$ & $32$ & $64$ & $64$ \\
\bottomrule
\end{tabular}
\end{table}

\paragraph{Parameter counts.}\label{app:lm-training-paramcounts}
We use the nominal model-size labels of GPT-3~\citep{brown_language_2020}. As shown in \Cref{tab:param-counts-lm}, IA and S4D-only have less than $1\%$ more parameters than Softmax. Most of the difference is the full complex readout $C_h$, which has $2HM^2$ real parameters per layer. Input normalization, SSM dynamics and short convolutions account for the rest.

\begin{table}[!htbp]
\centering
\caption{Trainable parameters of the language models. Counts include the untied input and output embeddings ($32{,}000$-token Llama 2 vocabulary~\citep{touvron_llama_2023-1}), the attention and SSM weights, the input RMSNorm scales and biases of \Cref{eq:inputnorm}, the layer RMSNorms and the SwiGLU layers.}
\label{tab:param-counts-lm}
\small
\begin{tabular}{l c c c c}
\toprule
Model & 125\,M & 350\,M & 760\,M & 1.3\,B \\
\midrule
Softmax     & $134{,}105{,}856$ & $373{,}867{,}520$ & $777{,}856{,}512$ & $1{,}345{,}423{,}360$ \\
S4D-only    & $135{,}425{,}424$ & $377{,}385{,}344$ & $781{,}523{,}328$ & $1{,}352{,}455{,}936$ \\
IA & $135{,}416{,}208$ & $377{,}360{,}768$ & $781{,}498{,}752$ & $1{,}352{,}406{,}784$ \\
\bottomrule
\end{tabular}
\end{table}

\paragraph{Optimizer and schedule.}\label{app:lm-training-optim}
The optimizer and schedule follow \citet{gu_linear-time_2024}. We use AdamW~\citep{loshchilov_decoupled_2019} with weight decay $0.1$ and gradient clipping at $1.0$. The peak learning rate is $3\times10^{-3}$ at 125\,M, $1.5\times10^{-3}$ at 350\,M, $1.25\times10^{-3}$ at 760\,M and $1\times10^{-3}$ at 1.3\,B. It rises linearly over 715 warmup updates (about 375\,M tokens) and then decays to $10^{-5}$ with a cosine schedule. Each update uses $524{,}288$ tokens in packed sequences.

The SSM parameters are not weight-decayed. The input-normalization parameters of \Cref{eq:inputnorm} and the dynamics parameters $\Delta_h$ and $A_h$ use a learning rate capped at $10^{-3}$, and the readout $C_h$ uses the main learning rate. Training uses bfloat16 mixed precision on 4 to 32 NVIDIA GH200 GPUs with distributed data parallelism (DDP). All pretraining and ablation runs use seed 42, and genomic fine-tuning uses seed 2222.

\clearpage
\section{Ablation settings and results}\label{app:adaptive-details}
\subsection{Ablation settings and parameter counts}
The LM configurations use 12 layers of width 768, 4 heads, feature/value width 192, 64 complex state modes and query-MLP hidden width 317. The genomic configurations use 4 layers of width 256, 16 heads, feature/value width 16, 8 complex modes and query hidden width 266. Q/K short convolutions have width 4 for language modeling and 8 for HG38, where the value convolution also has width 8. The query MLP output projection is initialized to zero. The first configuration, which has no Q/K short convolution, keeps the Q/K SiLU activations.

Each ablation configuration is trained once with the same seed on NVIDIA GH200 GPUs. Language-model configurations are trained for 4,768 updates of 524,288 tokens at context length 4096, and HG38 configurations for 10,000 updates of 1,024 sequences at context length 1024. Every configuration is validated on the same 80 batches of 8 sequences for language modeling, and on the same 100 batches of 128 sequences with 15\% masking for HG38.

\begin{table}[!htbp]
\centering\small
\setlength{\tabcolsep}{4pt}
\caption{Incremental ablations: final validation perplexity and parameter counts. Each row adds one component to the row above. $\Delta$NLL is the change from the previous row with a result. The value convolution is used only for HG38, so its language-modeling row is empty. Within each domain, all rows have the same recurrent-state size, and shared parameters are identically initialized.}
\label{tab:adaptive-ladder}
\label{tab:adaptive-parameters}
\begin{tabular}{@{}lrrr@{}}
\toprule
\multicolumn{4}{l}{\textit{Language modeling}} \\
\addlinespace[2pt]
Cumulative addition & Parameters & PPL $\downarrow$ & $\Delta$NLL $\downarrow$ \\
\midrule
Base IA without Q/K conv. & 134,528,304 & 16.595 & --- \\
Base IA (with Q/K conv.) & 134,620,464 & 16.361 & -0.0142 \\
+ Value conv.\ + SiLU & --- & --- & --- \\
+ Residual query MLP & 140,463,408 & 16.244 & -0.0071 \\
+ Value skip connection & 140,472,624 & 16.240 & -0.0002 \\
+ Output gate & 147,550,512 & 16.135 & -0.0065 \\
+ Selective writing & 147,578,160 & 16.095 & -0.0025 \\
+ Selective memory decay & 147,596,640 & 16.008 & -0.0054 \\
\midrule
\multicolumn{4}{l}{\textit{HG38 masked language modeling}} \\
\addlinespace[2pt]
Cumulative addition & Parameters & PPL $\downarrow$ & $\Delta$NLL $\downarrow$ \\
\midrule
Base IA without Q/K conv. & 1,076,864 & 2.737 & --- \\
Base IA (with Q/K conv.) & 1,113,728 & 2.689 & -0.0177 \\
+ Value conv.\ + SiLU & 1,122,944 & 2.661 & -0.0108 \\
+ Residual query MLP & 1,667,712 & 2.627 & -0.0126 \\
+ Value skip connection & 1,668,736 & 2.626 & -0.0006 \\
+ Output gate & 1,930,880 & 2.616 & -0.0036 \\
+ Selective writing & 1,933,952 & 2.616 & -0.0001 \\
+ Selective memory decay & 1,936,064 & 2.614 & -0.0008 \\
\bottomrule
\end{tabular}
\end{table}

\paragraph{Input processing in the mechanism ablation.}\label{app:mechanism-inputs}
With the key feature/value input, the key passes through a short convolution, SiLU and L2 normalization, and the value is not convolved. With the linear-projection input, both projections pass through a short convolution and SiLU without L2 normalization. Both inputs then use the per-head RMSNorm and bias of \Cref{eq:inputnorm}. The query-conditioned readout applies a short convolution, SiLU and L2 normalization to the query, and the linear readout has no query path. The value convolution and SiLU of \Cref{sec:adaptive} are not used.

\subsection{Native Caduceus baseline settings}\label{app:native-dna-settings}
The native Caduceus-PS baseline has 4 layers of width 256, expansion factor 2, state dimension 16 and 1,934,592 parameters, with bidirectional weight tying and reverse-complement parameter sharing. It is trained and validated like the Caduceus-IA ablation: HG38 at context length 1024, 1024 sequences per update, 10,000 updates and the validation batches above. Both use BF16 with TF32 disabled, peak learning rate 0.008, 1,000 warmup updates, cosine decay to 0.0001 and weight decay 0.1. The native model keeps its original initialization and BF16 residual accumulation. Caduceus-IA accumulates residuals in FP32.

\subsection{Caduceus-IA mechanism ablation on Base IA}\label{app:dna-mechanism-minimal}
\Cref{tab:dna-mechanism-minimal} repeats the Caduceus-IA mechanism ablation on Base IA, that is, without the residual query MLP, value convolution, value skip connection and output gate of \Cref{sec:adaptive}.

\begin{table}[!htbp]
\centering\footnotesize
\setlength{\tabcolsep}{4pt}
\caption{Caduceus-IA mechanism ablation on Base IA: final validation NLL and parameter counts at 1K context after 10,000 updates. Selective uses both selective writing and selective memory decay. Non-selective uses neither.}
\label{tab:dna-mechanism-minimal}
\label{tab:dna-mechanism-params}
\begin{tabular}{@{}llrrrr@{}}
\toprule
& & \multicolumn{2}{c}{Non-selective} & \multicolumn{2}{c}{Selective} \\
\cmidrule(lr){3-4}\cmidrule(lr){5-6}
Input & Readout & NLL $\downarrow$ & Parameters & NLL $\downarrow$ & Parameters \\
\midrule
Key feature/value & Query-conditioned & 0.9903 & 1,113,728 & 0.9896 & 1,118,912 \\
Key feature/value & Linear & 1.0099 & 1,096,320 & 0.9991 & 1,101,504 \\
Linear projection & Query-conditioned & 0.9792 & 1,132,160 & 0.9796 & 1,137,344 \\
Linear projection & Linear & 0.9847 & 1,114,752 & 0.9806 & 1,119,936 \\
\bottomrule
\end{tabular}
\end{table}

\clearpage
\section{Language-modeling scaling and downstream details}\label{app:lm-scaling-details}

\paragraph{Setup.}
We pretrain Llama-style decoder-only models (\appref{app:arch-details}) at four scales (125\,M, 350\,M, 760\,M, 1.3\,B parameters) on the FineWeb-Edu corpus~\citep{penedo_fineweb_2024} with the Llama 2 tokenizer~\citep{touvron_llama_2023-1} (32K vocabulary) and a training context length of $4096$. Each scale is trained at its Chinchilla-optimal token budget~\citep{hoffmann_empirical_2022} (approximately $20\times$ tokens per parameter): $2.5$\,B tokens at 125\,M, $7$\,B at 350\,M, $15$\,B at 760\,M and $26$\,B at 1.3\,B. Optimizer, schedule and hardware are given in \appref{app:lm-training-details}. We report the lowest validation perplexity reached during each run.

We compare Softmax, IA and S4D-only at all four scales, and S4D-only + RoPE at 125\,M. All models share the backbone, data, tokenizer and training recipe, and differ only in the sequence-mixing layer.
\begin{itemize}
\item \emph{Softmax}: multi-head softmax attention with RoPE.
\item \emph{IA}: the mechanism of \Cref{sec:architecture}.
\item \emph{S4D-only}: the S4D recurrence of IA with the full readout $C_h$, but with the linear-projection input $[a_t, b_t] = [W_a i_t, W_b i_t]$ and a linear readout (\Cref{fig:mechanism-arch}(d)).
\item \emph{S4D-only + RoPE}: S4D-only with RoPE applied to $a_t$.
\end{itemize}
IA and S4D-only have the same recurrent-state size at every scale (\appref{app:lm-state-budget}). Further 125\,M ablations of the two IA components and of RoPE are in \appref{app:mechanism-decomp}.

\paragraph{Scaling.}
\begin{table}[!htbp]
\centering\small
\caption{Language-modeling scaling: best-of-run validation PPL ($\downarrow$) on FineWeb-Edu. Parameter counts closely match across models at each nominal scale (\Cref{tab:param-counts-lm}).}
\label{tab:lm-scaling}
\begin{tabular}{lcccc}
\toprule
Nominal size & 125M & 350M & 760M & 1.3B \\
Training tokens & 2.5B & 7B & 15B & 26B \\
\midrule
Softmax & 16.65 & 12.12 & 10.09 & 8.62 \\
S4D-only & 19.22 & 13.51 & 10.67 & 9.13 \\
IA & $\mathbf{16.48}$ & $\mathbf{11.47}$ & $\mathbf{8.93}$ & $\mathbf{7.98}$ \\
\bottomrule
\end{tabular}
\end{table}

IA has $13$--$16\%$ lower validation perplexity than S4D-only at every scale. Relative to Softmax, IA is about $1\%$ lower at 125\,M and $7.4\%$ lower at 1.3\,B.

\begin{figure*}[!htbp]
\centering
\begin{tikzpicture}
\begin{axis}[
    width=0.6\linewidth,
    height=0.27\linewidth,
    xlabel={Training compute $6PD$ (FLOPs)},
    ylabel={Validation perplexity},
    xmode=log,
    ymode=log,
    log basis x={10},
    log basis y={10},
    xtick={1e18, 1e19, 1e20, 1e21},
    xticklabels={$10^{18}$, $10^{19}$, $10^{20}$, $10^{21}$},
    ytick={8, 10, 12, 16, 20},
    yticklabels={$8$, $10$, $12$, $16$, $20$},
    xmin=1.5e18, xmax=4e20,
    ymin=7.5, ymax=21,
    grid=major,
    grid style={dashed, gray!25},
    legend pos=north east,
    legend cell align=left,
    legend style={font=\footnotesize, draw=none, fill=none, row sep=-2pt},
    label style={font=\footnotesize},
    tick label style={font=\footnotesize},
    every axis plot/.append style={line width=1pt, mark size=2pt},
]
\addplot[color=IAplotGray, mark=square*, dashed] coordinates {
    (2.011e18, 16.65) (1.570e19, 12.12) (7.001e19, 10.09) (2.098e20, 8.62)
};
\addlegendentry{Softmax}
\addplot[color=IAplotBlue, mark=*, solid, line width=1.3pt] coordinates {
    (2.031e18, 16.48) (1.585e19, 11.47) (7.034e19, 8.93) (2.109e20, 7.98)
};
\addlegendentry{IA}
\addplot[color=IAplotOrange, mark=triangle*, dashdotted] coordinates {
    (2.031e18, 19.22) (1.585e19, 13.51) (7.034e19, 10.67) (2.109e20, 9.13)
};
\addlegendentry{S4D-only}
\end{axis}
\end{tikzpicture}
\caption{The runs of \Cref{fig:main-lm-scaling-context}(a) plotted against training compute. Compute is $6PD$ FLOPs for $P$ total parameters and $D$ training tokens~\citep{hoffmann_empirical_2022}. Both axes are logarithmic.}
\label{fig:lm-scaling}
\end{figure*}

\paragraph{Downstream evaluation at 1.3\,B.}
We evaluate the 1.3\,B Softmax, S4D-only, and IA models with the EleutherAI Language Model Evaluation Harness~\citep{biderman_lessons_2026} on the eight commonsense tasks of \citet{yang_gated_2025} (LAMBADA~\citep{paperno_lambada_2016}, PIQA~\citep{bisk_piqa_2020}, HellaSwag~\citep{zellers_hellaswag_2019}, WinoGrande~\citep{sakaguchi_winogrande_2020}, ARC-e/ARC-c~\citep{clark_think_2018}, SIQA~\citep{sap_social_2019}, BoolQ~\citep{clark_boolq_2019}), together with LAMBADA and WikiText-2~\citep{merity_pointer_2017} language-modeling perplexities. Aggregate and per-task results are in Appendices~\ref{app:lm-downstream} and~\ref{app:lm-commonsense-breakdown}. Relative to Softmax, IA raises the commonsense average by $3.03$ points and lowers bits per byte (BPB) by $0.0105$ on WikiText-2 and $0.0764$ on LAMBADA. S4D-only is $1.27$ points below Softmax on commonsense, and its LAMBADA perplexity is about twice as high.

\paragraph{Length extrapolation.}
We evaluate the 1.3\,B models at $N \in \{4\text{K}, 8\text{K}, 14\text{K}\}$ on six long-context corpora: PG19~\citep{rae_compressive_2020}, CodeParrot~\citep{tunstall_natural_2022}, GovReport~\citep{huang_efficient_2021}, NarrativeQA~\citep{kocisky_narrativeqa_2018}, Qasper~\citep{dasigi_dataset_2021}, and QMSum~\citep{zhong_qmsum_2021}. Within the 4\,K training context, Softmax has lower average perplexity than IA. Softmax perplexity rises to $22.85$ at 8\,K and $62.06$ at 14\,K. IA stays between $15.56$ and $15.77$, and S4D-only also changes little with length (\Cref{tab:lm-length-extrap}).

\begin{table}[!htbp]
\centering
\caption{Length-extrapolation perplexity at context length $N$, averaged over the six long-context corpora. Models are trained at 4\,K. Per-corpus results are in \appref{app:lm-length-extrap}.}
\label{tab:lm-length-extrap}
\small
\setlength{\tabcolsep}{4pt}
\begin{tabular}{l c c c}
\toprule
Context length & 4\,K (train) & 8\,K & 14\,K \\
\midrule
Softmax              & $\mathbf{14.21}$  & $22.85$           & $62.06$           \\
S4D-only           & $28.84$           & $28.48$           & $28.34$           \\
IA & $15.77$           & $\mathbf{15.56}$  & $\mathbf{15.71}$  \\
\bottomrule
\end{tabular}
\end{table}

\paragraph{Recall-heavy tasks.}
We evaluate recall at 1.3\,B on RULER single-needle-in-a-haystack~\citep{hsieh_ruler_2024}, Phonebook retrieval~\citep{jelassi_repeat_2024} and the Based zero-shot recall suite~\citep{arora_simple_2024} (\appref{app:lm-recall}). Within the 4\,K training context, Softmax is the most accurate in every setting except Based SQuAD. IA is more accurate than S4D-only on RULER and all six Based tasks, but both score zero on Phonebook. On LongBench-14~\citep{bai_longbench_2024}, IA is close to Softmax and well above S4D-only (\appref{app:lm-longbench}).

\clearpage
\section{Supplementary language-modeling results}\label{app:lm-supp}

\subsection{Mechanism and RoPE ablations at 125\,M}\label{app:mechanism-decomp}

We ablate the two components of IA, the key feature/value input $z_t = [\xi(k_t), v_t]$ and the query-conditioned readout $\xi(q_t)\,U_t^\top\,\Gamma_t$, together with RoPE. \Cref{tab:mechanism-decomp} reports six configurations and Softmax, and \Cref{fig:mechanism-arch} shows the four input and readout combinations they use.

\begin{table}[!htbp]
\centering
\caption{Mechanism and RoPE ablations at 125\,M / 2.5\,B tokens. ``vs Softmax'' is the relative change in validation perplexity, positive when worse than Softmax. Val.\ perplexity is the lowest of each run.}
\label{tab:mechanism-decomp}
\small
\setlength{\tabcolsep}{4pt}
\begin{tabularx}{\linewidth}{@{}
    >{\raggedright\arraybackslash}X
    r r@{}}
\toprule
Model & Val.\ perplexity & vs Softmax \\
\midrule
IA
& $\mathbf{16.48}$ & $\mathbf{-1.02\%}$ \\

IA, no RoPE
& $16.51$ & $-0.84\%$ \\

\addlinespace[2pt]
Softmax
& $16.65$ & N/A \\

\addlinespace[2pt]
Linear-projection input + query-conditioned readout
& $16.94$ & $+1.78\%$ \\

S4D-only
& $19.22$ & $+15.48\%$ \\

S4D-only + RoPE
& $19.77$ & $+18.79\%$ \\

Key feature/value input + linear readout
& $20.18$ & $+21.23\%$ \\
\bottomrule
\end{tabularx}
\end{table}

At the same recurrent-state size, IA has $14.3\%$ lower validation perplexity than S4D-only. Replacing the key feature/value input with the linear-projection input leaves perplexity slightly above Softmax, whereas removing the query-conditioned readout makes it worse than S4D-only. RoPE has little effect on IA and makes S4D-only worse.
\label{app:mechanism-arch}

\begin{figure}[!t]
\centering
\definecolor{mechA}{HTML}{2563EB}
\definecolor{mechB}{HTML}{A855F7}
\definecolor{mechC}{HTML}{F59E0B}
\definecolor{mechD}{HTML}{16A34A}
\definecolor{mechHeader}{HTML}{1E293B}
\definecolor{mechSsmFill}{HTML}{FDE68A}
\definecolor{mechSsmEdge}{HTML}{B45309}
\definecolor{mechIoFill}{HTML}{E2E8F0}
\definecolor{mechMergeFill}{HTML}{F1F5F9}
\definecolor{mechBypass}{HTML}{DC2626}

\tikzset{
    mechpic/.style={
        >=latex,
        x=.01\linewidth,
        y=.024\linewidth
    },
    nodebase/.style={
        rounded corners=1.2pt,
        minimum height=2.8mm,
        align=center,
        inner sep=0.55pt,
        font=\tiny
    },
    op/.style={
        nodebase,
        draw=#1,
        fill=#1!5,
        text width=7.1mm
    },
    op/.default={mechHeader},
    io/.style={
        nodebase,
        draw=mechHeader,
        fill=mechIoFill,
        font=\tiny\bfseries,
        text width=6.4mm
    },
    merge/.style={
        nodebase,
        draw=#1,
        fill=mechMergeFill,
        font=\tiny\bfseries,
        minimum height=3.5mm,
        inner ysep=0.85pt,
        text width=14.8mm
    },
    merge/.default={mechHeader},
    ssm/.style={
        nodebase,
        draw=mechSsmEdge,
        fill=mechSsmFill,
        line width=0.55pt,
        font=\tiny\bfseries,
        minimum height=5.6mm,
        inner ysep=1.5pt,
        text width=18.2mm
    },
    read/.style={
        nodebase,
        draw=#1,
        fill=#1!4,
        font=\tiny\bfseries,
        minimum height=6.0mm,
        inner ysep=1.5pt,
        text width=19.8mm
    },
    read/.default={mechHeader},
    flow/.style={
        ->,
        line width=0.45pt,
        draw=black!50,
        shorten <=0.5pt,
        shorten >=0.6pt
    },
    bypass/.style={
        ->,
        line width=0.62pt,
        densely dashed,
        draw=mechBypass,
        shorten <=0.5pt,
        shorten >=0.6pt
    }
}

\captionsetup[subfigure]{labelformat=parens,labelsep=space,justification=centering}

\begin{subfigure}[t]{.235\linewidth}
\centering
\begin{tikzpicture}[mechpic]
\path[use as bounding box] (0,0) rectangle (100,78);

\node[io] (A_x) at (50,73.5) {$x$};
\node[op=mechA] (A_wq) at (22,66.6) {$W_Q$};
\node[op=mechA] (A_wk) at (50,66.6) {$W_K$};
\node[op=mechA] (A_wv) at (78,66.6) {$W_V$};
\draw[flow, draw=mechA] (A_x.south) -- (A_wq.north);
\draw[flow, draw=mechA] (A_x.south) -- (A_wk.north);
\draw[flow, draw=mechA] (A_x.south) -- (A_wv.north);

\node[op=mechA] (A_scq) at (22,59.7) {SC};
\node[op=mechA] (A_sck) at (50,59.7) {SC};
\draw[flow, draw=mechA] (A_wq.south) -- (A_scq.north);
\draw[flow, draw=mechA] (A_wk.south) -- (A_sck.north);

\node[op=mechA, dashed] (A_rq) at (22,52.8) {RoPE};
\node[op=mechA, dashed] (A_rk) at (50,52.8) {RoPE};
\draw[flow, draw=mechA] (A_scq.south) -- (A_rq.north);
\draw[flow, draw=mechA] (A_sck.south) -- (A_rk.north);

\node[op=mechA] (A_xiq) at (22,45.9) {$\xi_q$};
\node[op=mechA] (A_xik) at (50,45.9) {$\xi_k$};
\node[op=mechA] (A_v) at (78,45.9) {$v$};
\draw[flow, draw=mechA] (A_rq.south) -- (A_xiq.north);
\draw[flow, draw=mechA] (A_rk.south) -- (A_xik.north);
\draw[flow, draw=mechA] (A_wv.south) -- (A_v.north);

\node[op=mechA] (A_nk) at (50,39.0) {N+b};
\node[op=mechA] (A_nv) at (78,39.0) {N+b};
\draw[flow, draw=mechA] (A_xik.south) -- (A_nk.north);
\draw[flow, draw=mechA] (A_v.south) -- (A_nv.north);

\node[merge=mechA] (A_u) at (64,31.8) {$[\xi_k,\,v]$};
\draw[flow, draw=mechA] (A_nk.south) -- (A_u.north);
\draw[flow, draw=mechA] (A_nv.south) -- (A_u.north);

\node[ssm] (A_ssm) at (50,23.0) {Complex S4D};
\draw[flow, draw=mechA] (A_u.south) -- (A_ssm.north);

\node[read=mechA] (A_read) at (50,12.5) {Query readout\\$\xi_q U^\top\Gamma$};
\draw[flow, draw=mechA] (A_ssm.south) -- (A_read.north);
\draw[bypass] (A_xiq.south west) .. controls (7,40.0) and (7,14.0) .. (A_read.west);

\node[io] (A_wo) at (50,3.6) {$W_O$};
\draw[flow, draw=mechA] (A_read.south) -- (A_wo.north);
\end{tikzpicture}
\subcaption{IA}
\end{subfigure}\hfill
\begin{subfigure}[t]{.235\linewidth}
\centering
\begin{tikzpicture}[mechpic]
\path[use as bounding box] (0,0) rectangle (100,78);

\node[io] (B_x) at (64,73.5) {$x$};
\node[op=mechB] (B_wk) at (50,66.6) {$W_K$};
\node[op=mechB] (B_wv) at (78,66.6) {$W_V$};
\draw[flow, draw=mechB] (B_x.south) -- (B_wk.north);
\draw[flow, draw=mechB] (B_x.south) -- (B_wv.north);

\node[op=mechB] (B_sck) at (50,59.7) {SC};
\draw[flow, draw=mechB] (B_wk.south) -- (B_sck.north);

\node[op=mechB] (B_rk) at (50,52.8) {RoPE};
\draw[flow, draw=mechB] (B_sck.south) -- (B_rk.north);

\node[op=mechB] (B_xik) at (50,45.9) {$\xi_k$};
\node[op=mechB] (B_v) at (78,45.9) {$v$};
\draw[flow, draw=mechB] (B_rk.south) -- (B_xik.north);
\draw[flow, draw=mechB] (B_wv.south) -- (B_v.north);

\node[op=mechB] (B_nk) at (50,39.0) {N+b};
\node[op=mechB] (B_nv) at (78,39.0) {N+b};
\draw[flow, draw=mechB] (B_xik.south) -- (B_nk.north);
\draw[flow, draw=mechB] (B_v.south) -- (B_nv.north);

\node[merge=mechB] (B_u) at (64,31.8) {$[\xi_k,\,v]$};
\draw[flow, draw=mechB] (B_nk.south) -- (B_u.north);
\draw[flow, draw=mechB] (B_nv.south) -- (B_u.north);

\node[ssm] (B_ssm) at (50,23.0) {Complex S4D};
\draw[flow, draw=mechB] (B_u.south) -- (B_ssm.north);

\node[read=mechB] (B_read) at (50,12.5) {Linear\\readout};
\draw[flow, draw=mechB] (B_ssm.south) -- (B_read.north);

\node[io] (B_wo) at (50,3.6) {$W_O$};
\draw[flow, draw=mechB] (B_read.south) -- (B_wo.north);
\end{tikzpicture}
\subcaption{Key feature/value input, linear readout}
\end{subfigure}\hfill
\begin{subfigure}[t]{.235\linewidth}
\centering
\begin{tikzpicture}[mechpic]
\path[use as bounding box] (0,0) rectangle (100,78);

\node[io] (C_x) at (50,73.5) {$x$};
\node[op=mechC] (C_wq) at (22,66.6) {$W_Q$};
\node[op=mechC] (C_wa) at (50,66.6) {$W_a$};
\node[op=mechC] (C_wb) at (78,66.6) {$W_b$};
\draw[flow, draw=mechC] (C_x.south) -- (C_wq.north);
\draw[flow, draw=mechC] (C_x.south) -- (C_wa.north);
\draw[flow, draw=mechC] (C_x.south) -- (C_wb.north);

\node[op=mechC] (C_scq) at (22,59.7) {SC};
\node[op=mechC] (C_sca) at (50,59.7) {SC};
\node[op=mechC] (C_scb) at (78,59.7) {SC};
\draw[flow, draw=mechC] (C_wq.south) -- (C_scq.north);
\draw[flow, draw=mechC] (C_wa.south) -- (C_sca.north);
\draw[flow, draw=mechC] (C_wb.south) -- (C_scb.north);

\node[op=mechC] (C_rq) at (22,52.8) {RoPE};
\node[op=mechC] (C_ra) at (50,52.8) {RoPE};
\draw[flow, draw=mechC] (C_scq.south) -- (C_rq.north);
\draw[flow, draw=mechC] (C_sca.south) -- (C_ra.north);

\node[op=mechC] (C_xiq) at (22,45.9) {$\xi_q$};
\node[op=mechC] (C_a) at (50,45.9) {$a$};
\node[op=mechC] (C_b) at (78,45.9) {$b$};
\draw[flow, draw=mechC] (C_rq.south) -- (C_xiq.north);
\draw[flow, draw=mechC] (C_ra.south) -- (C_a.north);
\draw[flow, draw=mechC] (C_scb.south) -- (C_b.north);

\node[op=mechC] (C_na) at (50,39.0) {N+b};
\node[op=mechC] (C_nb) at (78,39.0) {N+b};
\draw[flow, draw=mechC] (C_a.south) -- (C_na.north);
\draw[flow, draw=mechC] (C_b.south) -- (C_nb.north);

\node[merge=mechC] (C_u) at (64,31.8) {$[a,\,b]$};
\draw[flow, draw=mechC] (C_na.south) -- (C_u.north);
\draw[flow, draw=mechC] (C_nb.south) -- (C_u.north);

\node[ssm] (C_ssm) at (50,23.0) {Complex S4D};
\draw[flow, draw=mechC] (C_u.south) -- (C_ssm.north);

\node[read=mechC] (C_read) at (50,12.5) {Query readout\\$\xi_q U^\top\Gamma$};
\draw[flow, draw=mechC] (C_ssm.south) -- (C_read.north);
\draw[bypass] (C_xiq.south west) .. controls (7,40.0) and (7,14.0) .. (C_read.west);

\node[io] (C_wo) at (50,3.6) {$W_O$};
\draw[flow, draw=mechC] (C_read.south) -- (C_wo.north);
\end{tikzpicture}
\subcaption{Linear-projection input, query-conditioned readout}
\end{subfigure}\hfill
\begin{subfigure}[t]{.235\linewidth}
\centering
\begin{tikzpicture}[mechpic]
\path[use as bounding box] (0,0) rectangle (100,78);

\node[io] (D_x) at (64,73.5) {$x$};
\node[op=mechD] (D_wa) at (50,66.6) {$W_a$};
\node[op=mechD] (D_wb) at (78,66.6) {$W_b$};
\draw[flow, draw=mechD] (D_x.south) -- (D_wa.north);
\draw[flow, draw=mechD] (D_x.south) -- (D_wb.north);

\node[op=mechD] (D_sca) at (50,59.7) {SC};
\node[op=mechD] (D_scb) at (78,59.7) {SC};
\draw[flow, draw=mechD] (D_wa.south) -- (D_sca.north);
\draw[flow, draw=mechD] (D_wb.south) -- (D_scb.north);

\node[op=mechD, dashed] (D_ra) at (50,52.8) {RoPE};
\draw[flow, draw=mechD, densely dashed] (D_sca.south) -- (D_ra.north);

\node[op=mechD] (D_a) at (50,45.9) {$a$};
\node[op=mechD] (D_b) at (78,45.9) {$b$};
\draw[flow, draw=mechD, densely dashed] (D_ra.south) -- (D_a.north);
\draw[flow, draw=mechD] (D_scb.south) -- (D_b.north);

\node[op=mechD] (D_na) at (50,39.0) {N+b};
\node[op=mechD] (D_nb) at (78,39.0) {N+b};
\draw[flow, draw=mechD] (D_a.south) -- (D_na.north);
\draw[flow, draw=mechD] (D_b.south) -- (D_nb.north);

\node[merge=mechD] (D_u) at (64,31.8) {$[a,\,b]$};
\draw[flow, draw=mechD] (D_na.south) -- (D_u.north);
\draw[flow, draw=mechD] (D_nb.south) -- (D_u.north);

\node[ssm] (D_ssm) at (50,23.0) {Complex S4D};
\draw[flow, draw=mechD] (D_u.south) -- (D_ssm.north);

\node[read=mechD] (D_read) at (50,12.5) {Linear\\readout};
\draw[flow, draw=mechD] (D_ssm.south) -- (D_read.north);

\node[io] (D_wo) at (50,3.6) {$W_O$};
\draw[flow, draw=mechD] (D_read.south) -- (D_wo.north);
\end{tikzpicture}
\subcaption{S4D-only}
\end{subfigure}
\caption{The four input and readout combinations in the mechanism ablation. \textbf{(a)} \emph{IA} uses the key feature/value input $z_t=[\xi(k_t),v_t]$ and the query-conditioned readout $\xi(q_t)U_t^\top\Gamma_t$. \textbf{(b)} keeps the key feature/value input and replaces the readout with a learned linear readout of the SSM coefficients. \textbf{(c)} keeps the query-conditioned readout and uses the linear-projection input $[a_t,b_t]$. \textbf{(d)} \emph{S4D-only} uses neither. The dashed RoPE boxes in (a) and (d) mark RoPE that is varied across rows of \Cref{tab:mechanism-decomp}. The dashed red curve carries the query features to the readout. SC is a short convolution, and N${+}$b is the per-head RMSNorm and bias of \Cref{eq:inputnorm}, applied separately to $\xi(k_t)$ and $v_t$, or to $a_t$ and $b_t$.}
\label{fig:mechanism-arch}
\end{figure}

\subsection{State size at 1.3\,B}\label{app:lm-state-budget}

\Cref{tab:lm-state-budget} summarizes the recurrent-state size of the evaluated layers at the 1.3\,B scale, counting a complex state of dimension $n$ as $2n$ real degrees of freedom (DoF), as in \citet[Prop.~2]{lahoti_mamba-3_2026}. The SSM in IA takes the input $z_t = [\xi(k_t), v_t]$ of width $d_\text{feat} + d_h$, and we use $d_\text{feat} = d_h$. Each head therefore has a state of $2(d_\text{feat} + d_h)M$ real DoF. S4D-only takes an $[a, b]$ input of the same width, so both models have the same state size. The Softmax KV cache grows linearly with sequence length~$N$.

\begin{table}[!htbp]
\centering
\caption{Recurrent-state size at 1.3\,B ($d{=}2048$, $24$ layers, $H{=}32$, $d_h{=}64$, $M{=}64$). S4D-only + RoPE has the same state size as S4D-only.}
\label{tab:lm-state-budget}
\small
\setlength{\tabcolsep}{4pt}
\begin{tabular}{l c c c}
\toprule
Model & Heads & DoF per head & DoF per layer \\
\midrule
Softmax              & N/A & KV cache (grows with $N$) & $4{,}096\,N$ \\
S4D-only           & $H = 32$ & $2(d_h{+}d_h)\,M = 16{,}384$ & $524{,}288$ \\
IA & $H = 32$ & $2(d_\text{feat}{+}d_h)\,M = 16{,}384$ & $524{,}288$ \\
\bottomrule
\end{tabular}
\end{table}

\subsection{Downstream evaluation at 1.3\,B}\label{app:lm-downstream}\label{app:lm-longbench}

\Cref{tab:lm-downstream} reports validation loss and downstream metrics for the three 1.3\,B models. LongBench~\citep{bai_longbench_2024} uses the 14 tasks of \citet{yang_gated_2025}.

\begin{table}[!htbp]
\centering
\caption{Validation loss and downstream metrics for the 1.3\,B / $26$\,B-token runs. LAMBADA~\citep{paperno_lambada_2016} and WikiText-2~\citep{merity_pointer_2017} are evaluated zero-shot with the EleutherAI Language Model Evaluation Harness~\citep{biderman_lessons_2026}. Accuracy is reported in percent; LongBench scores are on $[0,1]$. BPB (bits per byte) is the total negative log-likelihood in bits divided by the number of UTF-8 bytes.}
\label{tab:lm-downstream}
\label{tab:lm-longbench}
\small
\begin{tabular}{l c c c}
\toprule
Metric & Softmax & S4D-only & IA \\
\midrule
Validation loss $\downarrow$ & $2.155$ & $2.212$ & $\mathbf{2.077}$ \\
Commonsense 8-task avg (\%) $\uparrow$ & $51.51$ & $50.23$ & $\mathbf{54.54}$ \\
LAMBADA accuracy (\%) $\uparrow$ & $39.43$ & $25.64$ & $\mathbf{44.36}$ \\
LAMBADA perplexity $\downarrow$ & $21.03$ & $41.02$ & $\mathbf{14.76}$ \\
LAMBADA BPB $\downarrow$ & $0.6578$ & $0.8021$ & $\mathbf{0.5814}$ \\
WikiText-2 perplexity $\downarrow$ & $20.55$ & $25.27$ & $\mathbf{19.77}$ \\
WikiText-2 BPB $\downarrow$ & $0.8156$ & $0.8713$ & $\mathbf{0.8051}$ \\
\midrule
LongBench 14-task avg $\uparrow$ & $\mathbf{0.1240}$ & $0.0857$ & $0.1229$ \\
\bottomrule
\end{tabular}
\end{table}

\subsection{Commonsense 8-task breakdown at 1.3\,B}\label{app:lm-commonsense-breakdown}

\begin{table}[!htbp]
\centering
\caption{Per-task accuracy (\%, higher is better) of the eight-task commonsense suite of \citet{yang_gated_2025}, evaluated on the 1.3\,B / $26$\,B-token runs with the EleutherAI Language Model Evaluation Harness~\citep{biderman_lessons_2026}. HellaSwag and ARC-challenge use length-normalized accuracy, and the other tasks use plain accuracy.}
\label{tab:lm-commonsense-breakdown}
\small
\begin{tabular}{l c c c}
\toprule
Task & Softmax & S4D-only & IA \\
\midrule
LAMBADA & $39.43$ & $25.64$ & $\mathbf{44.36}$ \\
PIQA & $69.42$ & $70.67$ & $\mathbf{71.38}$ \\
HellaSwag & $48.53$ & $49.34$ & $\mathbf{54.27}$ \\
WinoGrande & $54.22$ & $53.83$ & $\mathbf{57.85}$ \\
ARC-easy & $65.87$ & $67.55$ & $\mathbf{69.91}$ \\
ARC-challenge & $33.02$ & $33.36$ & $\mathbf{37.46}$ \\
SIQA & $40.28$ & $41.10$ & $\mathbf{41.45}$ \\
BoolQ & $\mathbf{61.28}$ & $60.37$ & $59.63$ \\
\midrule
\textbf{8-task avg} & $51.51$ & $50.23$ & $\mathbf{54.54}$ \\
\bottomrule
\end{tabular}
\end{table}
\subsection{Per-corpus length-extrapolation perplexity}\label{app:lm-length-extrap}

\begin{table}[!htbp]
\centering
\caption{Per-corpus validation perplexity at context lengths $N\in\{4,8,14\}$\,K for the 1.3\,B Softmax, S4D-only and IA models on PG19~\citep{rae_compressive_2020}, CodeParrot~\citep{tunstall_natural_2022}, GovReport~\citep{huang_efficient_2021}, NarrativeQA~\citep{kocisky_narrativeqa_2018}, Qasper~\citep{dasigi_dataset_2021}, and QMSum~\citep{zhong_qmsum_2021}. Training context is 4\,K, so $8$\,K and $14$\,K test length extrapolation. \Cref{tab:lm-length-extrap} reports the six-corpus average.}
\label{tab:lm-length-extrap-per-corpus}
\small
\setlength{\tabcolsep}{4pt}
\begin{tabular}{l l c c c}
\toprule
Corpus & Model & 4\,K & 8\,K & 14\,K \\
\midrule
\multirow{3}{*}{PG19}        & Softmax        & $16.34$  & $26.78$           & $70.95$           \\
                             & S4D-only     & $19.96$           & $19.75$           & $19.66$           \\
                             & IA & $\mathbf{15.86}$           & $\mathbf{15.73}$  & $\mathbf{15.94}$  \\
\midrule
\multirow{3}{*}{CodeParrot}  & Softmax        & $\mathbf{12.13}$  & $22.23$           & $74.85$           \\
                             & S4D-only     & $53.10$           & $52.44$           & $52.16$           \\
                             & IA & $17.62$           & $\mathbf{17.31}$  & $\mathbf{17.48}$  \\
\midrule
\multirow{3}{*}{GovReport}   & Softmax        & $\mathbf{7.33}$   & $12.20$           & $31.76$           \\
                             & S4D-only     & $9.19$            & $9.07$            & $9.01$            \\
                             & IA & $7.36$            & $\mathbf{7.27}$   & $\mathbf{7.29}$   \\
\midrule
\multirow{3}{*}{NarrativeQA} & Softmax        & $\mathbf{21.28}$  & $34.36$           & $92.75$           \\
                             & S4D-only     & $36.83$           & $36.42$           & $36.26$           \\
                             & IA & $22.85$           & $\mathbf{22.67}$  & $\mathbf{22.99}$  \\
\midrule
\multirow{3}{*}{Qasper}      & Softmax        & $\mathbf{14.43}$  & $22.94$           & $60.08$           \\
                             & S4D-only     & $23.53$           & $23.22$           & $23.13$           \\
                             & IA & $15.75$           & $\mathbf{15.52}$  & $\mathbf{15.63}$  \\
\midrule
\multirow{3}{*}{QMSum}       & Softmax        & $\mathbf{13.74}$  & $18.57$           & $41.96$           \\
                             & S4D-only     & $30.44$           & $29.97$           & $29.80$           \\
                             & IA & $15.20$           & $\mathbf{14.88}$  & $\mathbf{14.94}$  \\
\bottomrule
\end{tabular}
\end{table}

\subsection{Recall-heavy tasks at 1.3\,B}\label{app:lm-recall}

\Cref{tab:lm-recall} reports generation accuracy and the log-likelihood of the reference answer (\appref{app:lm-recall-likelihood}).

\begin{table}[!htbp]
\centering\footnotesize
\setlength{\tabcolsep}{4pt}
\caption{Recall evaluation of the 1.3\,B models: generation accuracy and teacher-forced reference-answer log-likelihood (nats per answer). RULER~\citep{hsieh_ruler_2024} uses S-NIAH-1; Phonebook~\citep{jelassi_repeat_2024} uses exact match; Based~\citep{arora_simple_2024} uses case-insensitive contains accuracy. Each synthetic setting has 200 examples; all examples in each Based validation split are scored for likelihood. $\Delta_{\mathrm{IA-S4D}}$ is the mean paired log-likelihood difference. Bold marks the highest value for each metric within each setting.}
\label{tab:lm-recall}
\label{tab:lm-recall-likelihood}
\begin{tabular}{@{}lrrrrrrr@{}}
\toprule
& \multicolumn{3}{c}{Accuracy (\%) $\uparrow$} & \multicolumn{3}{c}{Answer log-likelihood $\uparrow$} & \\
\cmidrule(lr){2-4}\cmidrule(lr){5-7}
Setting & Softmax & S4D-only & IA & Softmax & S4D-only & IA & $\Delta_{\mathrm{IA-S4D}}$ \\
\midrule
\multicolumn{8}{l}{\textit{RULER S-NIAH-1 (passkey retrieval)}} \\
1K context & $\mathbf{99.50}$ & $0.00$ & $65.50$ & $-2.14$ & $-18.16$ & $\mathbf{-1.52}$ & $+16.64$ \\
2K context & $\mathbf{100.00}$ & $0.00$ & $28.00$ & $\mathbf{-2.64}$ & $-20.24$ & $-7.60$ & $+12.65$ \\
4K context & $\mathbf{89.00}$ & $0.00$ & $9.50$ & $\mathbf{-3.46}$ & $-21.25$ & $-15.29$ & $+5.96$ \\
8K context & $0.00$ & $0.00$ & $\mathbf{3.50}$ & $-29.02$ & $-21.67$ & $\mathbf{-18.65}$ & $+3.02$ \\
\midrule
\multicolumn{8}{l}{\textit{Phonebook (exact match)}} \\
16 entries & $\mathbf{54.50}$ & $0.00$ & $0.00$ & $\mathbf{-1.81}$ & $-23.16$ & $-20.26$ & $+2.90$ \\
32 entries & $\mathbf{52.00}$ & $0.00$ & $0.00$ & $\mathbf{-2.25}$ & $-23.18$ & $-21.97$ & $+1.21$ \\
64 entries & $\mathbf{45.50}$ & $0.00$ & $0.00$ & $\mathbf{-3.62}$ & $-23.12$ & $-22.52$ & $+0.60$ \\
128 entries & $\mathbf{24.00}$ & $0.00$ & $0.00$ & $\mathbf{-7.06}$ & $-23.11$ & $-22.60$ & $+0.51$ \\
\midrule
\multicolumn{8}{l}{\textit{Based (case-insensitive contains accuracy)}} \\
SWDE & $\mathbf{60.58}$ & $6.57$ & $17.46$ & $\mathbf{-8.62}$ & $-15.95$ & $-13.74$ & $+2.21$ \\
FDA & $\mathbf{50.45}$ & $0.36$ & $9.53$ & $\mathbf{-12.10}$ & $-51.60$ & $-36.77$ & $+14.83$ \\
SQuAD & $11.33$ & $14.91$ & $\mathbf{33.78}$ & $-18.61$ & $-18.79$ & $\mathbf{-15.06}$ & $+3.73$ \\
NQ & $\mathbf{16.44}$ & $7.95$ & $11.43$ & $\mathbf{-9.75}$ & $-17.11$ & $-15.27$ & $+1.85$ \\
TriviaQA & $\mathbf{56.93}$ & $42.83$ & $54.27$ & $\mathbf{-9.54}$ & $-12.03$ & $-9.87$ & $+2.15$ \\
DROP & $\mathbf{23.47}$ & $14.04$ & $21.31$ & $\mathbf{-5.54}$ & $-8.44$ & $-6.07$ & $+2.37$ \\
Based average & $\mathbf{36.53}$ & $14.45$ & $24.63$ & --- & --- & --- & --- \\
\bottomrule
\end{tabular}
\end{table}

\subsection{Reference-answer likelihood on recall-heavy tasks}
\label{app:lm-recall-likelihood}
We score the reference answer under teacher forcing with the pretrained 1.3\,B models, without fine-tuning. For a prompt $x$ and reference answer $y$, the score is
\begin{equation}
\ell_\theta(x,y)=\sum_{t=1}^{|y|}\log p_\theta(y_t\mid x,y_{<t}),
\end{equation}
where the sum covers the answer tokens, including the answer delimiter and excluding EOS. Prompts are preprocessed and truncated as in the accuracy evaluation, and all models score the same tokens. For examples with several references, we use the first. We report the mean log-likelihood per example in nats and the mean paired difference between IA and S4D-only.

IA has a higher mean log-likelihood than S4D-only in all 14 settings. On Phonebook, both have zero accuracy, but IA is ahead by $0.51$ to $2.90$ nats per answer.

\clearpage
\section{Genomic modeling details}\label{app:genomics-supp}

\subsection{Pretraining}
\begin{table}[!htbp]
\centering\small
\caption{HG38 masked language modeling at 1K context after 10,000 updates. Both models have width 256. Caduceus-IA-PS is the last row of \Cref{tab:adaptive-ladder}, and Caduceus-PS is the native baseline of \appref{app:native-dna-settings}.}
\label{tab:dna-pretraining}
\begin{tabular}{lcc}
\toprule
Model & Parameters & Validation PPL $\downarrow$ \\
\midrule
Caduceus-PS & 1.935M & 2.612 \\
Caduceus-IA-PS & 1.936M & 2.614 \\
\bottomrule
\end{tabular}
\end{table}

\subsection{Genomic downstream results}\label{app:dna-downstream}
Baseline results are taken from \citet{schiff_caduceus_2024}, and we follow their evaluation protocol. GenomicBenchmarks results are over 5 split seeds and Nucleotide Transformer results over 10.

For GenomicBenchmarks, the model width is 118 and the IA layers have 16 heads of width 7, an internal width of 112. The model has 470,254 parameters. We chose this head layout over $7\times16$ and $8\times14$, which have the same internal width, by pretraining validation loss. For Nucleotide Transformer, we use the width-256 model of \Cref{tab:dna-pretraining}.

Fine-tuning updates the full backbone with mean pooling and no head dropout. It uses weight decay 0.1, gradient clipping at 1.0 and BF16 mixed precision with TF32 disabled. The learning rate warms up over the first 1\% of training and decays to 0.1 of its peak. Caduceus-IA-Ph is fine-tuned with reverse-complement augmentation, following Section~4.2 of \citet{schiff_caduceus_2024}. Caduceus-IA-PS is fine-tuned without it.

On GenomicBenchmarks, both variants train for 10 epochs with effective batch size 256. Caduceus-IA-PS uses learning rate 0.001. Caduceus-IA-Ph uses the learning rates reported for Caduceus-Ph by \citet[Tables~5--6]{schiff_caduceus_2024}.

On Nucleotide Transformer, both variants train for 20 epochs, and we report the test score of the epoch with the best validation metric. Caduceus-IA-PS uses learning rate 0.001, or 0.002 for Enhancer types. Its effective batch size is 512 for H3K4me2, H3K4me3, Enhancers, Enhancer types, Promoter TATA and Splice sites all, and 128 for the other 12 tasks. Caduceus-IA-Ph uses the learning rates and batch sizes reported for Caduceus-Ph in the same tables.

\begin{table}[!htbp]
\centering\small
\renewcommand{\arraystretch}{1.1}
\setlength{\tabcolsep}{2pt}
\caption{GenomicBenchmarks accuracy. Mean $\pm$ range (maximum minus minimum) over five split seeds. Baselines are from Table~1 of \citet{schiff_caduceus_2024}. Caduceus-IA uses the width-118 model. Cad.\ is Caduceus and IA is Caduceus-IA. Each cell shows the mean above the range, and bold marks the highest mean in each row.}
\label{tab:gb}
\begin{tabular}{@{}>{\raggedright\arraybackslash}p{2.7cm}cccccccc@{}}
\toprule
Task & CNN & HyenaDNA & Mamba & \shortstack{Cad.\\w/o equiv.} & \shortstack{Cad.\\Ph} & \shortstack{Cad.\\PS} & \shortstack{IA\\Ph} & \shortstack{IA\\PS} \\
\midrule
Mouse enhancers & \shortstack{$0.715$\\$\pm 0.087$} & \shortstack{$0.780$\\$\pm 0.025$} & \shortstack{$0.743$\\$\pm 0.054$} & \shortstack{$0.770$\\$\pm 0.058$} & \shortstack{$0.754$\\$\pm 0.074$} & \shortstack{$\mathbf{0.793}$\\$\mathbf{\pm 0.058}$} & \shortstack{$0.751$\\$\pm 0.050$} & \shortstack{$0.775$\\$\pm 0.066$} \\
Coding vs. intergenomic & \shortstack{$0.892$\\$\pm 0.008$} & \shortstack{$0.904$\\$\pm 0.005$} & \shortstack{$0.904$\\$\pm 0.004$} & \shortstack{$0.908$\\$\pm 0.003$} & \shortstack{$0.915$\\$\pm 0.003$} & \shortstack{$0.910$\\$\pm 0.003$} & \shortstack{$\mathbf{0.933}$\\$\mathbf{\pm 0.002}$} & \shortstack{$0.931$\\$\pm 0.005$} \\
Human vs. worm & \shortstack{$0.942$\\$\pm 0.002$} & \shortstack{$0.964$\\$\pm 0.002$} & \shortstack{$0.967$\\$\pm 0.002$} & \shortstack{$0.970$\\$\pm 0.003$} & \shortstack{$\mathbf{0.973}$\\$\mathbf{\pm 0.001}$} & \shortstack{$0.968$\\$\pm 0.002$} & \shortstack{$\mathbf{0.973}$\\$\mathbf{\pm 0.001}$} & \shortstack{$0.969$\\$\pm 0.002$} \\
Human enhancers Cohn & \shortstack{$0.702$\\$\pm 0.021$} & \shortstack{$0.729$\\$\pm 0.014$} & \shortstack{$0.732$\\$\pm 0.029$} & \shortstack{$0.741$\\$\pm 0.008$} & \shortstack{$\mathbf{0.747}$\\$\mathbf{\pm 0.004}$} & \shortstack{$0.745$\\$\pm 0.007$} & \shortstack{$0.745$\\$\pm 0.013$} & \shortstack{$0.737$\\$\pm 0.006$} \\
Human enhancer Ensembl & \shortstack{$0.744$\\$\pm 0.122$} & \shortstack{$0.849$\\$\pm 0.006$} & \shortstack{$0.862$\\$\pm 0.008$} & \shortstack{$0.883$\\$\pm 0.002$} & \shortstack{$0.893$\\$\pm 0.008$} & \shortstack{$\mathbf{0.900}$\\$\mathbf{\pm 0.006}$} & \shortstack{$\mathbf{0.900}$\\$\mathbf{\pm 0.002}$} & \shortstack{$\mathbf{0.900}$\\$\mathbf{\pm 0.009}$} \\
Human regulatory & \shortstack{$0.872$\\$\pm 0.005$} & \shortstack{$0.869$\\$\pm 0.012$} & \shortstack{$0.814$\\$\pm 0.211$} & \shortstack{$0.871$\\$\pm 0.007$} & \shortstack{$0.872$\\$\pm 0.011$} & \shortstack{$0.873$\\$\pm 0.007$} & \shortstack{$\mathbf{0.888}$\\$\mathbf{\pm 0.002}$} & \shortstack{$0.881$\\$\pm 0.002$} \\
Human OCR Ensembl & \shortstack{$0.698$\\$\pm 0.013$} & \shortstack{$0.783$\\$\pm 0.007$} & \shortstack{$0.815$\\$\pm 0.002$} & \shortstack{$0.818$\\$\pm 0.003$} & \shortstack{$\mathbf{0.828}$\\$\mathbf{\pm 0.006}$} & \shortstack{$0.818$\\$\pm 0.006$} & \shortstack{$0.822$\\$\pm 0.004$} & \shortstack{$0.792$\\$\pm 0.009$} \\
Human non-TATA promoters & \shortstack{$0.861$\\$\pm 0.009$} & \shortstack{$0.944$\\$\pm 0.002$} & \shortstack{$0.933$\\$\pm 0.007$} & \shortstack{$0.933$\\$\pm 0.006$} & \shortstack{$0.946$\\$\pm 0.007$} & \shortstack{$0.945$\\$\pm 0.010$} & \shortstack{$\mathbf{0.954}$\\$\mathbf{\pm 0.007}$} & \shortstack{$0.950$\\$\pm 0.004$} \\
\bottomrule
\end{tabular}
\end{table}

\begin{table}[!htbp]
\centering\footnotesize
\renewcommand{\arraystretch}{1.1}
\setlength{\tabcolsep}{2pt}
\caption{Nucleotide Transformer downstream tasks. Mean $\pm$ range over ten split seeds. Baselines are from Table~2 of \citet{schiff_caduceus_2024}. Notation as in \Cref{tab:gb}.}
\label{tab:nt}
\begin{tabular}{@{}>{\raggedright\arraybackslash}p{2.15cm}lcccccccc@{}}
\toprule
Task & Metric & Enformer & DNABERT-2 & NT-v2 & HyenaDNA & \shortstack{Cad.\\Ph} & \shortstack{Cad.\\PS} & \shortstack{IA\\Ph} & \shortstack{IA\\PS} \\
\midrule
H3 & MCC & \shortstack{$0.719$\\$\pm 0.048$} & \shortstack{$0.785$\\$\pm 0.033$} & \shortstack{$0.784$\\$\pm 0.047$} & \shortstack{$0.779$\\$\pm 0.037$} & \shortstack{$0.815$\\$\pm 0.048$} & \shortstack{$0.799$\\$\pm 0.029$} & \shortstack{$\mathbf{0.827}$\\$\mathbf{\pm 0.032}$} & \shortstack{$0.811$\\$\pm 0.032$} \\
H3K14ac & MCC & \shortstack{$0.288$\\$\pm 0.077$} & \shortstack{$0.516$\\$\pm 0.028$} & \shortstack{$0.551$\\$\pm 0.021$} & \shortstack{$0.612$\\$\pm 0.065$} & \shortstack{$0.631$\\$\pm 0.026$} & \shortstack{$0.541$\\$\pm 0.212$} & \shortstack{$\mathbf{0.652}$\\$\mathbf{\pm 0.022}$} & \shortstack{$0.625$\\$\pm 0.051$} \\
H3K36me3 & MCC & \shortstack{$0.344$\\$\pm 0.055$} & \shortstack{$0.591$\\$\pm 0.020$} & \shortstack{$0.625$\\$\pm 0.013$} & \shortstack{$0.613$\\$\pm 0.041$} & \shortstack{$0.601$\\$\pm 0.129$} & \shortstack{$0.609$\\$\pm 0.109$} & \shortstack{$\mathbf{0.658}$\\$\mathbf{\pm 0.024}$} & \shortstack{$0.641$\\$\pm 0.035$} \\
H3K4me1 & MCC & \shortstack{$0.291$\\$\pm 0.061$} & \shortstack{$0.511$\\$\pm 0.028$} & \shortstack{$\mathbf{0.550}$\\$\mathbf{\pm 0.021}$} & \shortstack{$0.512$\\$\pm 0.024$} & \shortstack{$0.523$\\$\pm 0.039$} & \shortstack{$0.488$\\$\pm 0.102$} & \shortstack{$0.546$\\$\pm 0.032$} & \shortstack{$0.509$\\$\pm 0.038$} \\
H3K4me2 & MCC & \shortstack{$0.211$\\$\pm 0.069$} & \shortstack{$0.336$\\$\pm 0.040$} & \shortstack{$0.319$\\$\pm 0.045$} & \shortstack{$0.455$\\$\pm 0.095$} & \shortstack{$0.487$\\$\pm 0.170$} & \shortstack{$0.388$\\$\pm 0.101$} & \shortstack{$\mathbf{0.520}$\\$\mathbf{\pm 0.059}$} & \shortstack{$0.490$\\$\pm 0.042$} \\
H3K4me3 & MCC & \shortstack{$0.158$\\$\pm 0.072$} & \shortstack{$0.352$\\$\pm 0.077$} & \shortstack{$0.410$\\$\pm 0.033$} & \shortstack{$0.549$\\$\pm 0.056$} & \shortstack{$0.544$\\$\pm 0.045$} & \shortstack{$0.440$\\$\pm 0.202$} & \shortstack{$\mathbf{0.571}$\\$\mathbf{\pm 0.033}$} & \shortstack{$0.521$\\$\pm 0.037$} \\
H3K79me3 & MCC & \shortstack{$0.496$\\$\pm 0.042$} & \shortstack{$0.613$\\$\pm 0.030$} & \shortstack{$0.626$\\$\pm 0.026$} & \shortstack{$0.672$\\$\pm 0.048$} & \shortstack{$0.697$\\$\pm 0.077$} & \shortstack{$0.676$\\$\pm 0.026$} & \shortstack{$\mathbf{0.716}$\\$\mathbf{\pm 0.015}$} & \shortstack{$0.698$\\$\pm 0.039$} \\
H3K9ac & MCC & \shortstack{$0.420$\\$\pm 0.063$} & \shortstack{$0.542$\\$\pm 0.029$} & \shortstack{$0.562$\\$\pm 0.040$} & \shortstack{$0.581$\\$\pm 0.061$} & \shortstack{$0.622$\\$\pm 0.030$} & \shortstack{$0.604$\\$\pm 0.048$} & \shortstack{$\mathbf{0.651}$\\$\mathbf{\pm 0.023}$} & \shortstack{$0.607$\\$\pm 0.041$} \\
H4 & MCC & \shortstack{$0.732$\\$\pm 0.076$} & \shortstack{$0.796$\\$\pm 0.027$} & \shortstack{$0.799$\\$\pm 0.025$} & \shortstack{$0.763$\\$\pm 0.044$} & \shortstack{$\mathbf{0.811}$\\$\mathbf{\pm 0.022}$} & \shortstack{$0.789$\\$\pm 0.020$} & \shortstack{$0.808$\\$\pm 0.028$} & \shortstack{$0.803$\\$\pm 0.025$} \\
H4ac & MCC & \shortstack{$0.273$\\$\pm 0.063$} & \shortstack{$0.463$\\$\pm 0.041$} & \shortstack{$0.495$\\$\pm 0.032$} & \shortstack{$0.564$\\$\pm 0.038$} & \shortstack{$0.621$\\$\pm 0.054$} & \shortstack{$0.525$\\$\pm 0.240$} & \shortstack{$\mathbf{0.627}$\\$\mathbf{\pm 0.058}$} & \shortstack{$0.593$\\$\pm 0.038$} \\
Enhancers & MCC & \shortstack{$0.451$\\$\pm 0.108$} & \shortstack{$0.516$\\$\pm 0.098$} & \shortstack{$\mathbf{0.548}$\\$\mathbf{\pm 0.144}$} & \shortstack{$0.517$\\$\pm 0.117$} & \shortstack{$0.546$\\$\pm 0.073$} & \shortstack{$0.491$\\$\pm 0.066$} & \shortstack{$0.513$\\$\pm 0.050$} & \shortstack{$0.522$\\$\pm 0.071$} \\
Enhancer types & MCC & \shortstack{$0.309$\\$\pm 0.134$} & \shortstack{$0.423$\\$\pm 0.051$} & \shortstack{$0.424$\\$\pm 0.132$} & \shortstack{$0.386$\\$\pm 0.185$} & \shortstack{$\mathbf{0.439}$\\$\mathbf{\pm 0.054}$} & \shortstack{$0.416$\\$\pm 0.095$} & \shortstack{$0.428$\\$\pm 0.069$} & \shortstack{$0.395$\\$\pm 0.115$} \\
Promoter all & F1 & \shortstack{$0.954$\\$\pm 0.006$} & \shortstack{$0.971$\\$\pm 0.006$} & \shortstack{$\mathbf{0.976}$\\$\mathbf{\pm 0.006}$} & \shortstack{$0.960$\\$\pm 0.005$} & \shortstack{$0.970$\\$\pm 0.004$} & \shortstack{$0.967$\\$\pm 0.004$} & \shortstack{$0.968$\\$\pm 0.003$} & \shortstack{$0.965$\\$\pm 0.006$} \\
Promoter no TATA & F1 & \shortstack{$0.955$\\$\pm 0.010$} & \shortstack{$0.972$\\$\pm 0.005$} & \shortstack{$\mathbf{0.976}$\\$\mathbf{\pm 0.005}$} & \shortstack{$0.959$\\$\pm 0.008$} & \shortstack{$0.969$\\$\pm 0.011$} & \shortstack{$0.968$\\$\pm 0.006$} & \shortstack{$0.969$\\$\pm 0.006$} & \shortstack{$0.966$\\$\pm 0.008$} \\
Promoter TATA & F1 & \shortstack{$0.960$\\$\pm 0.023$} & \shortstack{$0.955$\\$\pm 0.021$} & \shortstack{$\mathbf{0.966}$\\$\mathbf{\pm 0.013}$} & \shortstack{$0.944$\\$\pm 0.040$} & \shortstack{$0.953$\\$\pm 0.016$} & \shortstack{$0.957$\\$\pm 0.015$} & \shortstack{$0.952$\\$\pm 0.021$} & \shortstack{$0.948$\\$\pm 0.021$} \\
Splice sites all & Acc. & \shortstack{$0.848$\\$\pm 0.019$} & \shortstack{$0.939$\\$\pm 0.009$} & \shortstack{$\mathbf{0.983}$\\$\mathbf{\pm 0.008}$} & \shortstack{$0.956$\\$\pm 0.011$} & \shortstack{$0.940$\\$\pm 0.027$} & \shortstack{$0.927$\\$\pm 0.021$} & \shortstack{$0.970$\\$\pm 0.005$} & \shortstack{$0.967$\\$\pm 0.008$} \\
Splice acceptor & F1 & \shortstack{$0.914$\\$\pm 0.028$} & \shortstack{$0.975$\\$\pm 0.006$} & \shortstack{$\mathbf{0.981}$\\$\mathbf{\pm 0.011}$} & \shortstack{$0.958$\\$\pm 0.010$} & \shortstack{$0.937$\\$\pm 0.033$} & \shortstack{$0.936$\\$\pm 0.077$} & \shortstack{$0.973$\\$\pm 0.007$} & \shortstack{$0.967$\\$\pm 0.010$} \\
Splice donor & F1 & \shortstack{$0.906$\\$\pm 0.027$} & \shortstack{$0.963$\\$\pm 0.006$} & \shortstack{$\mathbf{0.985}$\\$\mathbf{\pm 0.022}$} & \shortstack{$0.949$\\$\pm 0.024$} & \shortstack{$0.948$\\$\pm 0.025$} & \shortstack{$0.874$\\$\pm 0.289$} & \shortstack{$0.976$\\$\pm 0.009$} & \shortstack{$0.968$\\$\pm 0.017$} \\
\bottomrule
\end{tabular}
\end{table}

\paragraph{Variant-effect prediction.}
We evaluate causal-eQTL prediction with frozen embeddings of the reference and alternative sequences. The embeddings come from a different Caduceus-IA model: width 256, 7.7M parameters, pretrained on HG38 at context length 131,072 for 50,000 updates (\Cref{sec:exp-genomics}). Its IA chunk length is 64 for Caduceus-IA-PS and 16 for Caduceus-IA-Ph. An RBF-kernel SVM predicts variant effects from these embeddings. Variants are grouped by distance to the nearest transcription start site: 0--30\,kb, 30--100\,kb, and over 100\,kb. For each group, we train five SVMs on independently sampled subsets of 5,000 training examples and evaluate on held-out chromosomes 9 and 10.

Caduceus-IA-PS matches Caduceus-PS below 30\,kb, is slightly better at 30--100\,kb and slightly worse beyond 100\,kb (\Cref{tab:vep}). Caduceus-IA-Ph has a lower mean AUROC than Caduceus-Ph in all three groups.

\begin{table}[!htbp]
\centering\small
\renewcommand{\arraystretch}{1.1}
\setlength{\tabcolsep}{2pt}
\caption{Causal-eQTL variant-effect prediction. Test AUROC by distance to the nearest transcription start site. Caduceus-IA cells show the mean above the sample standard deviation over five SVM fits. Baseline means are from Figure~4 of \citet{schiff_caduceus_2024}. Notation as in \Cref{tab:gb}.}
\label{tab:vep}
\begin{tabular}{@{}>{\raggedright\arraybackslash}p{2.7cm}cccccccc@{}}
\toprule
Distance & Enformer & NT-v2 & HyenaDNA &
\shortstack{Cad.\\w/o equiv.} &
\shortstack{Cad.\\Ph} & \shortstack{Cad.\\PS} & \shortstack{IA\\Ph} & \shortstack{IA\\PS} \\
\midrule
0--30\,kb
& 0.668 & 0.671 & 0.670 & 0.674 & 0.678 & $\mathbf{0.679}$
& \shortstack{$0.673$\\$\pm 0.003$} & \shortstack{$\mathbf{0.679}$\\$\mathbf{\pm 0.005}$} \\
30--100\,kb
& $\mathbf{0.659}$ & 0.593 & 0.599 & 0.623 & 0.648 & 0.612
& \shortstack{$0.639$\\$\pm 0.018$} & \shortstack{$0.620$\\$\pm 0.018$} \\
$>100$\,kb
& 0.563 & 0.540 & 0.569 & 0.562 & 0.580 & $\mathbf{0.586}$
& \shortstack{$0.562$\\$\pm 0.010$} & \shortstack{$0.574$\\$\pm 0.009$} \\
\bottomrule
\end{tabular}
\end{table}

\clearpage
\section{Inference scaling}\label{app:inference-scaling}

\subsection{Decode latency versus prefix length}
\label{app:cuda-graph-decode}

All inference measurements use one NVIDIA RTX 6000 Ada Generation GPU (48\,GB) with BF16 mixed precision, PyTorch 2.10, CUDA 12.8 and no PyTorch compilation (\texttt{torch.compile}), with randomly initialized weights. This subsection uses the 1.3\,B architecture of \appref{app:lm-training-backbone}. Softmax uses PyTorch SDPA with the Flash backend and a preallocated KV cache of $N+64$ tokens, without CUDA graphs. IA is run with and without CUDA graphs. In both cases it uses chunked prefill with chunk size 2048 and contracts the query before the SSM output projection when decoding.

\paragraph{Measurement.}\label{app:decode-protocol}
Each model runs in a separate process. After warmup, we run prefill followed by 64 decode steps timed with CUDA events, and repeat this 20 times (15 at $B{=}32$ and 10 at $B{=}64$). The latency at a prefix length is the median over decode steps 2--64 of all repetitions. Speedup is the median Softmax latency divided by the median latency of IA with CUDA graphs, both medians taken over the prefix lengths that both models complete.

\begin{figure}[!htbp]
\centering
{\scriptsize
\begin{tabular}{@{}c@{\hspace{1.4em}}c@{\hspace{1.4em}}c@{}}
\tikz[baseline=-0.5ex]{\draw[IAplotGray,thick] (0,0) -- (0.55,0); \fill[IAplotGray] (0.275,0) rectangle ++(0.045,0.045);} Softmax (SDPA) &
\tikz[baseline=-0.5ex]{\draw[IAplotBlue,thick,dashed] (0,0) -- (0.55,0); \fill[IAplotBlue] (0.275,0.045) -- (0.235,-0.035) -- (0.315,-0.035) -- cycle;} IA (CUDA graph) &
\tikz[baseline=-0.5ex]{\draw[IAplotBlue!55!white,thick,densely dotted] (0,0) -- (0.55,0); \draw[IAplotBlue!55!white,thick] (0.275,0) circle (0.035);} IA (no CUDA graph)
\end{tabular}
}
\vspace{0.1cm}

\begin{subfigure}[t]{0.48\textwidth}
\centering
\begin{tikzpicture}
\begin{axis}[
  width=\textwidth, height=4.5cm,
  xlabel={Prefix length}, ylabel={ms / step},
  xmode=log, log basis x=2,
  xtick={512,1024,2048,4096,8192,16384},
  xticklabels={512,1k,2k,4k,8k,16k},
  ymin=0, ymax=45,
  title={\small $B = 1$},
  grid=major, grid style={gray!30},
]
\addplot[IAplotGray,thick,mark=square*,mark size=1.5pt] coordinates {
  (512,13.6694) (1024,14.0478) (2048,13.7820) (4096,12.8097) (8192,12.8952) (16384,14.0933)
};
\addplot[IAplotBlue,thick,dashed,mark=triangle*,mark size=1.5pt] coordinates {
  (512,12.2972) (1024,12.3136) (2048,12.3228) (4096,12.2982) (8192,12.3331) (16384,12.3197)
};
\addplot[IAplotBlue!55!white,thick,densely dotted,mark=o,mark size=1.5pt] coordinates {
  (512,31.3605) (1024,30.7092) (2048,32.2452) (4096,29.9167) (8192,31.8935) (16384,32.1178)
};
\end{axis}
\end{tikzpicture}
\end{subfigure}
\hfill
\begin{subfigure}[t]{0.48\textwidth}
\centering
\begin{tikzpicture}
\begin{axis}[
  width=\textwidth, height=4.5cm,
  xlabel={Prefix length}, ylabel={ms / step},
  xmode=log, log basis x=2,
  xtick={512,1024,2048,4096,8192,16384},
  xticklabels={512,1k,2k,4k,8k,16k},
  ymin=0, ymax=45,
  title={\small $B = 8$},
  grid=major, grid style={gray!30},
]
\addplot[IAplotGray,thick,mark=square*,mark size=1.5pt] coordinates {
  (512,13.1003) (1024,13.5956) (2048,13.3673) (4096,12.9802) (8192,20.5282)
};
\addplot[IAplotBlue,thick,dashed,mark=triangle*,mark size=1.5pt] coordinates {
  (512,14.1486) (1024,14.1650) (2048,14.1752) (4096,14.1742) (8192,14.1957) (16384,14.1947)
};
\addplot[IAplotBlue!55!white,thick,densely dotted,mark=o,mark size=1.5pt] coordinates {
  (512,30.8946) (1024,31.4235) (2048,31.8925) (4096,33.9697) (8192,33.3256) (16384,33.0982)
};
\end{axis}
\end{tikzpicture}
\end{subfigure}

\vspace{0.3cm}

\begin{subfigure}[t]{0.48\textwidth}
\centering
\begin{tikzpicture}
\begin{axis}[
  width=\textwidth, height=4.5cm,
  xlabel={Prefix length}, ylabel={ms / step},
  xmode=log, log basis x=2,
  xtick={512,1024,2048,4096,8192,16384},
  xticklabels={512,1k,2k,4k,8k,16k},
  ymin=0, ymax=45,
  title={\small $B = 16$},
  grid=major, grid style={gray!30},
]
\addplot[IAplotGray,thick,mark=square*,mark size=1.5pt] coordinates {
  (512,14.2131) (1024,13.0959) (2048,14.0247) (4096,21.0616)
};
\addplot[IAplotBlue,thick,dashed,mark=triangle*,mark size=1.5pt] coordinates {
  (512,15.6918) (1024,15.6887) (2048,15.7082) (4096,15.7481) (8192,15.7696) (16384,15.7870)
};
\addplot[IAplotBlue!55!white,thick,densely dotted,mark=o,mark size=1.5pt] coordinates {
  (512,32.6682) (1024,31.6836) (2048,32.6569) (4096,33.0225) (8192,34.0383) (16384,40.3809)
};
\end{axis}
\end{tikzpicture}
\end{subfigure}
\hfill
\begin{subfigure}[t]{0.48\textwidth}
\centering
\begin{tikzpicture}
\begin{axis}[
  width=\textwidth, height=4.5cm,
  xlabel={Prefix length}, ylabel={ms / step},
  xmode=log, log basis x=2,
  xtick={512,1024,2048,4096,8192,16384},
  xticklabels={512,1k,2k,4k,8k,16k},
  ymin=0, ymax=45,
  title={\small $B = 32$},
  grid=major, grid style={gray!30},
]
\addplot[IAplotGray,thick,mark=square*,mark size=1.5pt] coordinates {
  (512,14.3186) (1024,13.9684) (2048,21.7641)
};
\addplot[IAplotBlue,thick,dashed,mark=triangle*,mark size=1.5pt] coordinates {
  (512,19.1345) (1024,19.1754) (2048,19.1980) (4096,19.2973) (8192,19.2399) (16384,19.3126)
};
\addplot[IAplotBlue!55!white,thick,densely dotted,mark=o,mark size=1.5pt] coordinates {
  (512,36.2271) (1024,41.5969) (2048,32.1915) (4096,34.0060) (8192,32.3523) (16384,33.0926)
};
\end{axis}
\end{tikzpicture}
\end{subfigure}

\caption{Decode latency of the 1.3\,B models versus prefix length. A curve ends at the largest prefix length the model completes without running out of memory (\Cref{tab:decode-summary}).}
\label{fig:decode-scaling}
\end{figure}

\begin{table}[!htbp]
\centering
\caption{Decode latency at 1.3\,B. Ranges are the minimum and maximum over completed prefix lengths, and ``Largest $N$'' is the largest completed prefix length. Speedup is defined in \appref{app:decode-protocol}.}
\label{tab:decode-summary}
\small
\setlength{\tabcolsep}{3pt}
\begin{tabular}{@{}r cc c ccc c@{}}
\toprule
& \multicolumn{2}{c}{Softmax (SDPA)} && \multicolumn{3}{c}{IA} & \\
\cmidrule(lr){2-3} \cmidrule(lr){5-7}
$B$ & Range (ms) & Largest $N$ && No graph (ms) & CUDA graph (ms) & Largest $N$ & Speedup\\
\midrule
1 & $12.81$--$14.09$ & $16{,}384$ && $29.92$--$32.25$ & $12.30$--$12.33$ & $16{,}384$ & $1.11\times$ \\
8 & $12.98$--$20.53$ & $8{,}192$ && $30.89$--$33.97$ & $14.15$--$14.20$ & $16{,}384$ & $0.94\times$ \\
16 & $13.10$--$21.06$ & $4{,}096$ && $31.68$--$40.38$ & $15.69$--$15.79$ & $16{,}384$ & $0.90\times$ \\
32 & $13.97$--$21.76$ & $2{,}048$ && $32.19$--$41.60$ & $19.13$--$19.31$ & $16{,}384$ & $0.75\times$ \\
64 & $24.10$--$29.06$ & $1{,}024$ && $64.93$--$66.64$ & $38.41$ & $1{,}024$ & $0.69\times$ \\
\bottomrule
\end{tabular}
\end{table}

\subsection{Prefill memory}
\label{app:prefill-memory}

\Cref{tab:prefill-memory} reports peak memory during prefill at 1.3\,B. Softmax and non-chunked IA run a full forward pass that includes the vocabulary logits. Chunked IA only builds the decoding state and computes no logits, so the columns are not directly comparable.

\begin{table}[!htbp]
\centering
\caption{Peak memory (GB) during prefill at 1.3\,B. Chunked IA uses chunk size 2048. OOM means out of memory.}
\label{tab:prefill-memory}
\small
\begin{tabular}{rrccc}
\toprule
$B$ & $N$ & Softmax (SDPA) & IA (non-chunked) & IA (chunked) \\
\midrule
8 & $4{,}096$ & $14.65$ & $24.42$ & $14.85$ \\
8 & $8{,}192$ & $21.42$ & $41.02$ & $15.39$ \\
8 & $16{,}384$ & $35.35$ & OOM & $12.38$ \\
16 & $4{,}096$ & $21.21$ & OOM & $21.71$ \\
32 & $2{,}048$ & $21.16$ & $40.95$ & $33.09$ \\
64 & $512$ & $14.59$ & $24.78$ & $26.70$ \\
\bottomrule
\end{tabular}
\end{table}

\subsection{Decode latency across model sizes}
\label{app:decode-scaling-sizes}

\Cref{tab:decode-scaling-sizes} gives decode latency at all four model sizes. At every size and at both $B{=}1$ and $B{=}8$, the latency of IA with CUDA graphs varies by less than $1\%$ across prefix lengths.

\begin{table}[!htbp]
\centering\small
\setlength{\tabcolsep}{4pt}
\caption{Decode latency across model sizes. Latencies are ranges over completed prefix lengths in $\{512,\ldots,16{,}384\}$. At $B=1$ every prefix length completes. At $B=8$, the largest completed prefix length is in parentheses, and the last column is the ratio of the largest to the smallest Softmax latency.}
\label{tab:decode-scaling-sizes}
\label{tab:decode-scaling-sizes-b8}
\begin{tabular}{@{}rcccc@{}}
\toprule
\multicolumn{5}{l}{\textit{Batch size $B=1$}} \\
\addlinespace[2pt]
Size & Softmax (ms) & IA, CUDA graph (ms) & Speedup & \\
\midrule
125\,M & $6.47$--$6.91$ & $2.37$--$2.38$ & $2.81\times$ & \\
350\,M & $12.89$--$14.14$ & $5.51$--$5.53$ & $2.46\times$ & \\
760\,M & $13.13$--$15.07$ & $8.45$--$8.48$ & $1.63\times$ & \\
1.3\,B & $12.81$--$14.09$ & $12.30$--$12.33$ & $1.11\times$ & \\
\midrule
\multicolumn{5}{l}{\textit{Batch size $B=8$}} \\
\addlinespace[2pt]
Size & Softmax (ms) & IA, CUDA graph (ms) & Speedup & Softmax max/min \\
\midrule
125\,M & $6.80$--$7.48$ ($16$\,K) & $2.81$--$2.83$ ($16$\,K) & $2.46\times$ & $1.10\times$ \\
350\,M & $13.38$--$16.97$ ($16$\,K) & $6.54$--$6.56$ ($16$\,K) & $2.10\times$ & $1.27\times$ \\
760\,M & $14.58$--$15.68$ ($8$\,K) & $9.86$--$9.92$ ($16$\,K) & $1.53\times$ & $1.08\times$ \\
1.3\,B & $12.98$--$20.53$ ($8$\,K) & $14.15$--$14.20$ ($16$\,K) & $0.94\times$ & $1.58\times$ \\
\bottomrule
\end{tabular}
\end{table}

\clearpage
\section{Training throughput on NVIDIA GH200}\label{app:training-speed}

We time a full optimizer update for the 1.3\,B language models of \Cref{sec:exp-new-lm} and for the width-256, four-layer genomic models of \Cref{tab:dna-pretraining}, with randomly initialized weights. IA uses chunk length 64. An IA update takes $2.82\times$ as long as a Softmax update and $1.92\times$ as long as an S4D-only update. A Caduceus-IA-PS update takes $3.25\times$ as long as a Caduceus-PS update (\Cref{tab:training-speed-gh200}).

\begin{table}[htbp]
\centering
\caption{Cost of one training update on four NVIDIA GH200 GPUs. Throughput is summed over the GPUs, and peak memory is per GPU.}
\label{tab:training-speed-gh200}
\small
\setlength{\tabcolsep}{4pt}
\begin{tabular}{@{}lrrrrr@{}}
\toprule
Model & Parameters & \shortstack{Median\\s/update} & \shortstack{IQR\\(s)} & \shortstack{Throughput\\(ktokens/s)} & \shortstack{Peak memory\\(GiB/GPU)} \\
\midrule
\multicolumn{6}{l}{\textit{Language modeling ($N=4096$, 524,288 tokens/update)}} \\
\addlinespace[2pt]
Softmax & 1345.4M & 3.598 & 0.005 & 145.7 & 48.77 \\
S4D-only & 1352.5M & 5.285 & 0.017 & 99.2 & 65.50 \\
IA & 1352.4M & 10.140 & 0.014 & 51.7 & 67.20 \\
\midrule
\multicolumn{6}{l}{\textit{HG38 masked language modeling ($N=1024$, 1,048,576 tokens/update)}} \\
\addlinespace[2pt]
Caduceus-PS & 1.935M & 0.297 & 0.001 & 3534.4 & 10.40 \\
Caduceus-IA-PS & 1.936M & 0.963 & 0.001 & 1088.4 & 58.48 \\
\bottomrule
\end{tabular}
\end{table}

\paragraph{Workloads.}
All five models run one after another on one node with four NVIDIA GH200 GPUs, in bfloat16 mixed precision. For language modeling, each GPU processes two sequences of length 4096 per microbatch and accumulates 16 microbatches, giving 524,288 tokens per optimizer update. For HG38 masked language modeling, each GPU processes 128 sequences of length 1024 and accumulates two microbatches, giving 1,024 sequences (1,048,576 input tokens) per update. Within each domain, all models see identical batches and MLM masks.

\paragraph{Implementation and precision.}
The language models are compiled with PyTorch's \texttt{torch.compile} and use TF32. The genomic models run without compilation and with TF32 disabled. Softmax uses PyTorch SDPA with the Flash backend, S4D-only uses an FFT convolution, IA uses Triton chunkwise kernels and Caduceus-IA-PS uses TileLang chunkwise kernels. Caduceus-PS uses its Mamba kernels. Caduceus-IA-PS accumulates residuals in FP32 and Caduceus-PS in BF16. Activation checkpointing is off.

\paragraph{Measurement.}
A timed update covers the forward and backward passes, gradient accumulation and communication, clipping and the optimizer step. It excludes data loading and validation. After five warmup updates, we time 20 updates in one run per model. The time of an update is the longest of the four GPUs, measured between CUDA synchronizations. We report the median and the interquartile range (IQR) of these 20 times. Throughput is the number of input tokens per update divided by the median time. Peak memory is the largest PyTorch allocation on any GPU after warmup.

The software environment is PyTorch 2.10.0 with CUDA 12.6, Triton 3.6.0, TileLang 0.1.14, Mamba-SSM 2.3.1, causal-conv1d 1.6.1, and NCCL 2.27.5.

\end{document}